\documentclass[10pt, conference, compsocconf]{IEEEtran}
\usepackage{ifpdf}
\usepackage{cite}
\usepackage[pdftex]{graphicx}
\usepackage[cmex10]{amsmath}

\usepackage{array}
\usepackage{mdwmath}
\usepackage{mdwtab}
\usepackage{eqparbox}
\usepackage[font=footnotesize]{subfig}
\usepackage{stfloats}
\usepackage{url}

\def\BibTeX{{\rm B\kern-.05em{\sc i\kern-.025em b}\kern-.08em
    T\kern-.1667em\lower.7ex\hbox{E}\kern-.125emX}}

\usepackage[nolist]{acronym}
\usepackage[switch,columnwise]{lineno}
\usepackage{booktabs}
\usepackage{amsmath,amssymb}
\usepackage{hyperref}
\usepackage{cleveref}
\usepackage{algorithm}
\usepackage[noend]{algpseudocode}
\usepackage[super]{nth}
\usepackage{csquotes}
\usepackage{wrapfig}
\usepackage{xcolor}
\usepackage[binary-units=true]{siunitx}

\usepackage{xspace}
\newcommand{\etal}{\textit{et~al}.~}
\newcommand{\eg}{e.g.,\xspace} 
\newcommand{\ie}{i.e.,\xspace} 
\newcommand{\cf}{c.f.\xspace}

\usepackage{todonotes}

\definecolor{darkgreen}{rgb}{0.0,0.5,0.0}
\definecolor{pastelgreen}{rgb}{0.42,0.72,0.65}
\definecolor{aquamarine}{rgb}{0.13, 0.42, 0.53}
\definecolor{orange}{rgb}{0.8,0.4,0}
\definecolor{purple}{rgb}{0.5,0,0.5}

\newcommand{\belowfig}{\vspace{-0.5cm}}
\newcommand{\beforesubsec}{\vspace{-0.0cm}}

\newcommand{\StyleTune}{\textit{StyleTune}\xspace}

\begin{document}
%
\title{Interactive Multi-level Stroke Control for Neural Style Transfer}


\author{
\IEEEauthorblockN{Max Reimann, Benito Buchheim}
\IEEEauthorblockA{Hasso Plattner Institute\\ 
University of Potsdam, Germany\\
max.reimann@hpi.de\\benito.buchheim@student.hpi.de}
\and
\IEEEauthorblockN{Amir Semmo}
\IEEEauthorblockA{Digital Masterpieces GmbH\\
Potsdam, Germany\\
amir.semmo@digitalmasterpieces.com}
\and
\IEEEauthorblockN{J\"urgen D\"ollner, Matthias Trapp}
\IEEEauthorblockA{Hasso Plattner Institute\\
University of Potsdam, Germany\\
juergen.doellner@hpi.de\\
matthias.trapp@hpi.de}
}



%

\newcommand{\specFigSize}{0.235\textwidth}
\newcommand{\halfFigSize}{0.15\textwidth}
\newlength{\tempdima}
\newcommand{\rowname}[1]
{\rotatebox{90}{\makebox[\tempdima][c]{\text{#1}}}}


\maketitle

\begin{abstract}
We present \StyleTune, a mobile app for interactive multi-level control of neural style transfers that facilitates creative adjustments of style elements and enables high output fidelity. 
In contrast to current mobile neural style transfer apps, \StyleTune supports users to adjust both the size and orientation of style elements, such as brushstrokes and texture patches, on a global as well as local level.
To this end, we propose a novel stroke-adaptive feed-forward style transfer network, that enables control over stroke size and intensity and allows a larger range of edits than current approaches. 
For additional level-of-control, we propose a network agnostic method for stroke-orientation adjustment by utilizing the rotation-variance of \acp{CNN}. 
To achieve high output fidelity, we further add a patch-based style transfer method that enables users to obtain output resolutions of more than 20 \ac{Mpix}. Our approach empowers users to create many novel results that are not possible with current mobile neural style transfer apps.
\end{abstract}


%
\IEEEpeerreviewmaketitle

\section{Introduction}
\label{ST:Sec:Introduction}

\noindent Machine learning has become of prior interest to both research and end-user applications of image-based artistic rendering~\cite{tewari2020state}. 
Its usage in mobile expressive rendering setups has particularly increased over the last years, providing essential tools for casual creativity and image filtering~\cite{amato2019ai}. Here, a popular method is to extract the style from one exemplar image and transfer it to a target image or video, thus making a generalized \ac{NST} practicable~\cite{jing2019neural}.


\acp{NST} enjoy a great popularity with both users and developers because of its ability to emulate artistic styles without the need to engineer style-specific algorithms. 
While \acp{NST} have also found their way into several professional tools, such as Photoshop \cite{DapkusNeuralFilter}, they are mostly limited to resembling \enquote{one click solutions}.
In particular, these implementations are typically constrained to pre-defined styles that can be applied globally to a target image, thus enabling higher-level interaction, but without lower-level control that is often seeked by artists \cite{Isenberg2016} and prosumers of image filtering apps \cite{KPSD17}. For instance, no control over perceptual elements of a style such as stroke placement or style granularity are provided, \ie inherently limiting the degree of artistic expression \cite{SID17}. 
Further, existing approaches for low-level control of \acp{NST} generally only considers a univariate adjustment of the stylized outputs, which makes a complex and individual editing impracticable. 
Furthermore, current mobile apps are quite limited in their output resolutions, which typically stems from inherent hardware limitations. 

\begin{figure*}[t!]
\centerline{
\subfloat[Content and style images]{%
\includegraphics[height=3.5cm]{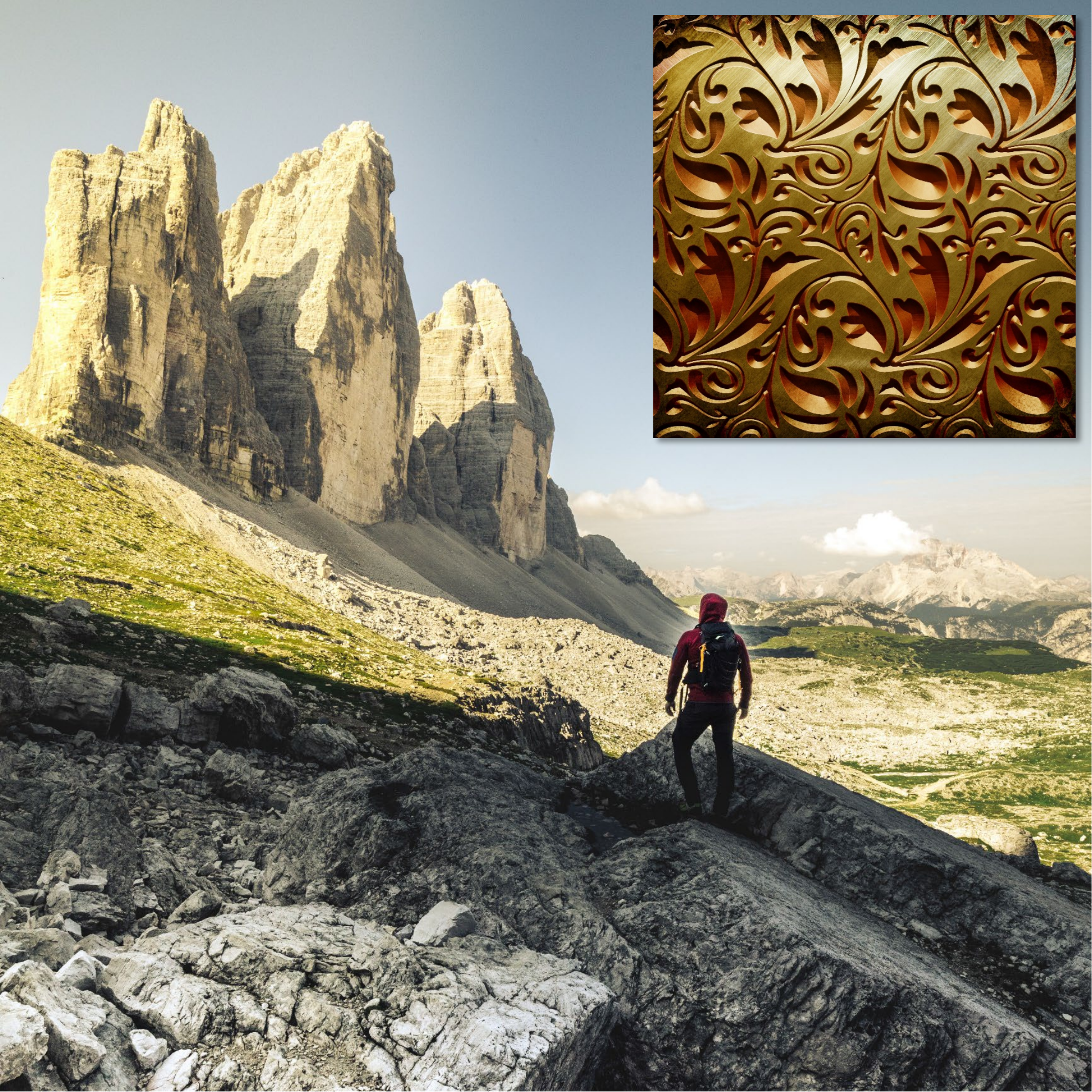}%
\label{ST:Fig:Teaser:A}%
}\hfill
\subfloat[Global transfer]{%
\includegraphics[height=3.5cm]{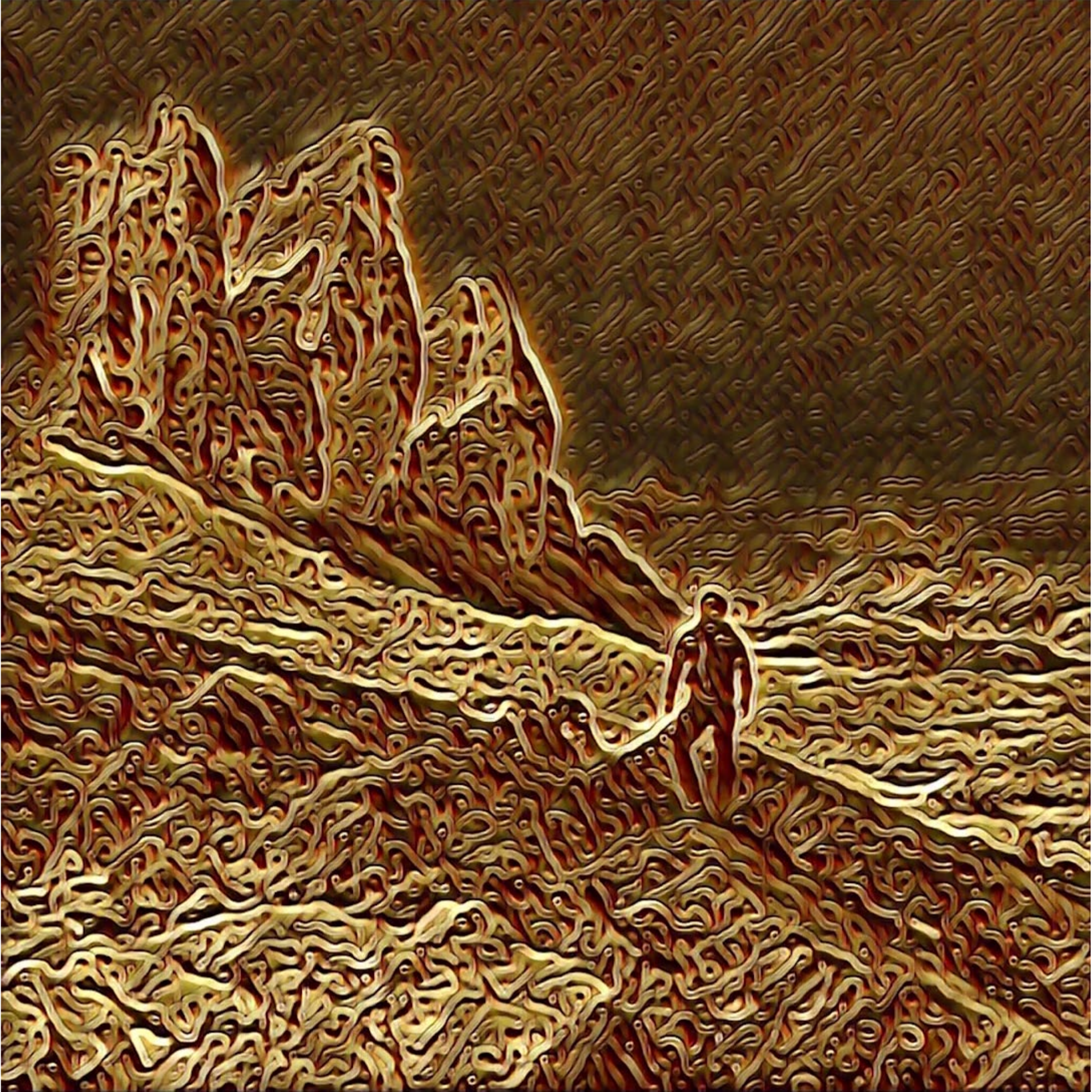}%
\label{ST:Fig:Teaser:B}%
}\hfill
\subfloat[Local adjustments]{%
\includegraphics[height=3.5cm]{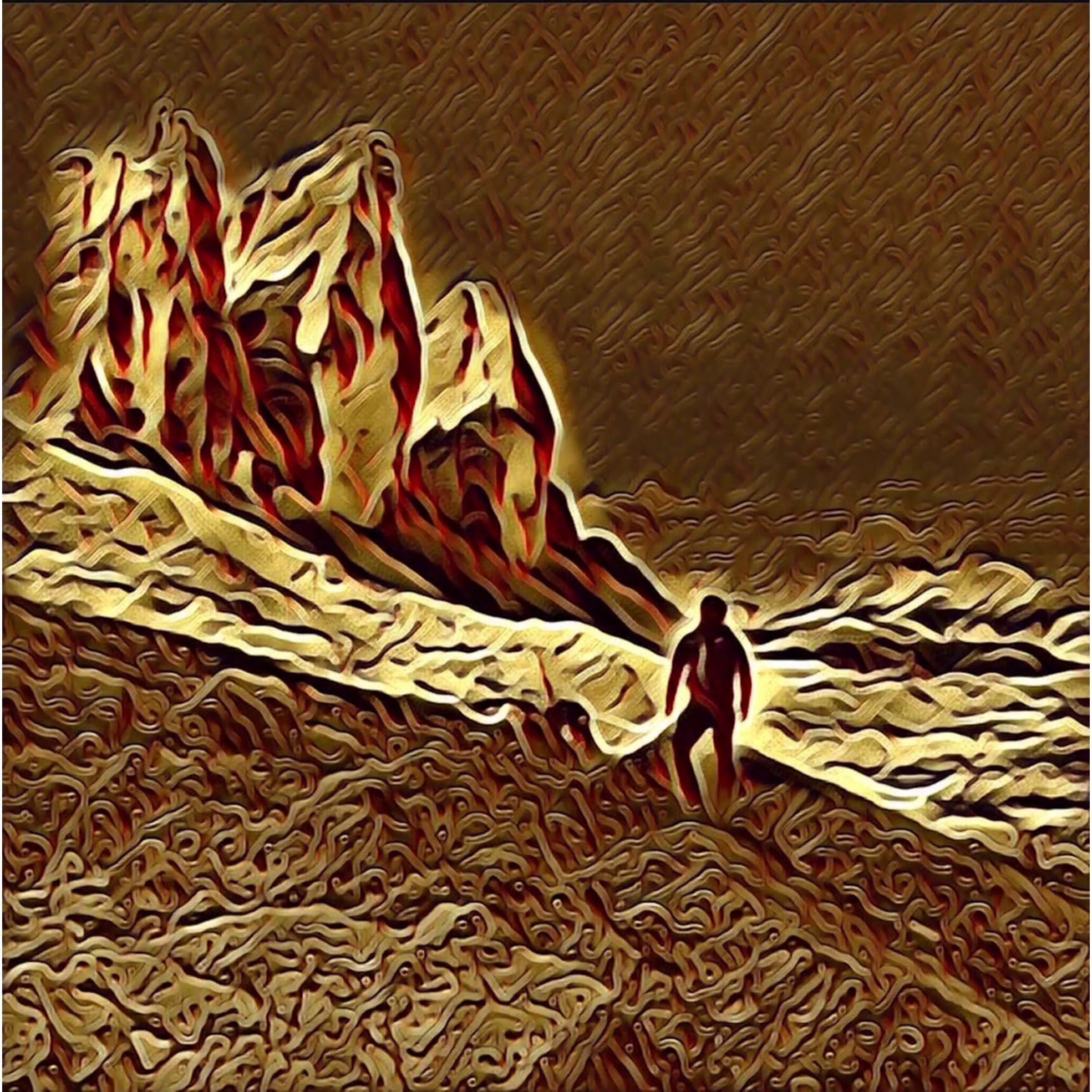}%
\label{ST:Fig:Teaser:C}%
}\hfill
\subfloat[Upsampled \ac{NST} and zoomed versions]{%
\includegraphics[height=3.5cm]{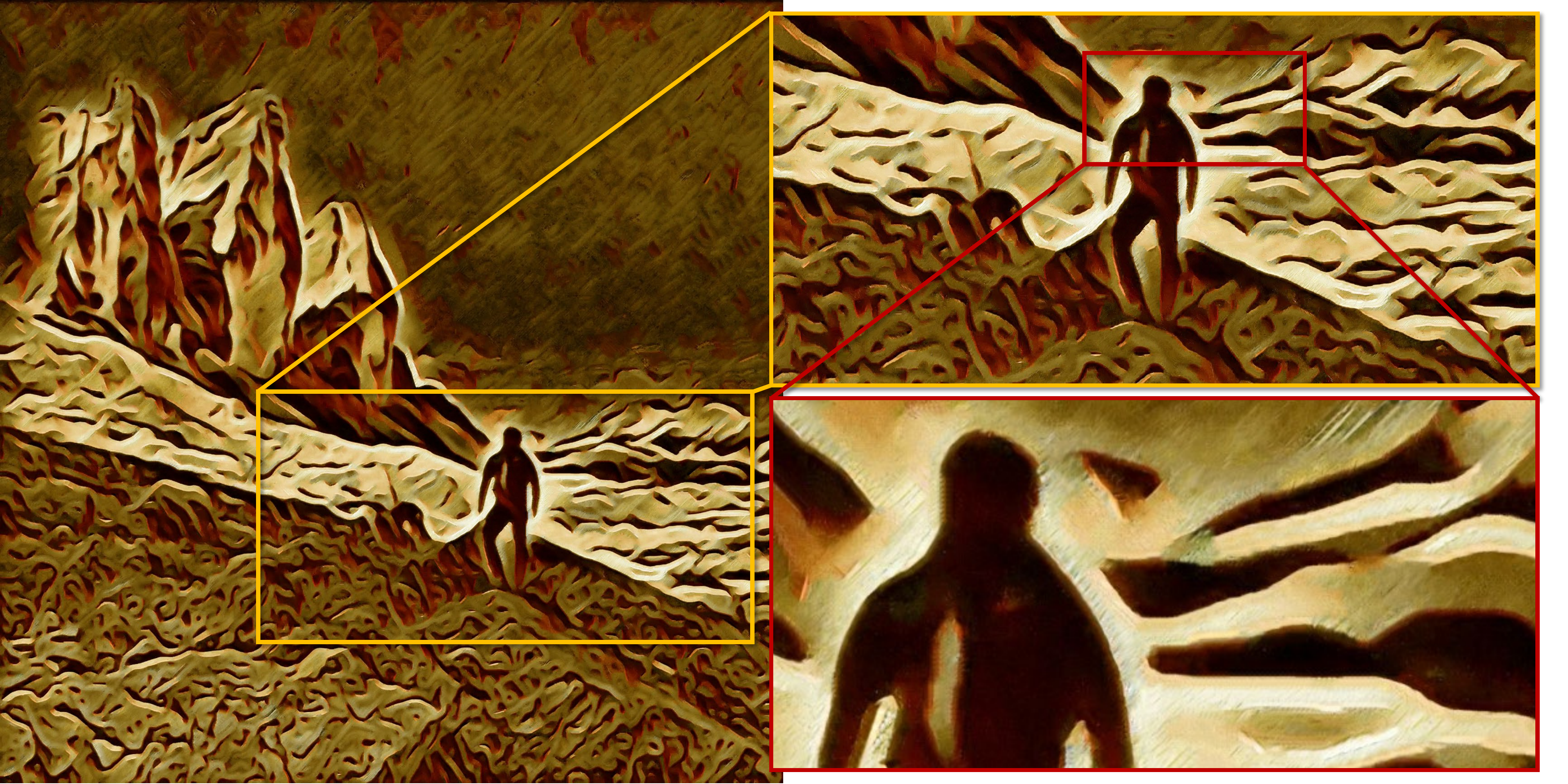}%
\label{ST:Fig:Teaser:D}%
}}
\caption{Comparison between a global style transfer \protect\subref{ST:Fig:Teaser:B} and a locally-retouched version produced with \StyleTune \protect\subref{ST:Fig:Teaser:C} based on given content and style images \protect\subref{ST:Fig:Teaser:A}. Figure \protect\subref{ST:Fig:Teaser:D} shows a high-resolution result obtained by style-guided upsampling of \protect\subref{ST:Fig:Teaser:C} from $1024 \times 1024$ pixels to a resolution of $3200 \times 3200$ pixels. (content image from \textcopyright{} Shutterstock, used with permission).}
\label{ST:Fig:Teaser}
\belowfig 
\end{figure*}

This work presents an approach for multivariate editing of \acp{NST} using a novel feedforward style transfer network that alleviates the aforementioned limitations by incorporating global and local control over style elements, such as their granularity and orientation. We demonstrate the real-world applicability of our concept by presenting a mobile app (\StyleTune) for interactive editing of \acp{NST}.
In addition, our app enables to create high-resolution outputs of the edited results to advance the field of mobile style transfer towards art-directable tools for both casual and professional creativity~(\Cref{ST:Fig:Teaser}).
To summarize, this paper makes the following contributions:
\begin{enumerate}
\item A method for level-of-control interaction to adjust a style's stroke size, intensity, and rotation with a single neural network.
\item A method for efficient upsampling and adaptive global control over \ac{NST} outputs using a novel two-stream network architecture.
\item An interactive editing pipeline for \acp{NST} on mobile devices, with the ability to achieve high-resolution outputs of \num{20} \ac{Mpix} and more.
\end{enumerate}

\noindent The remainder of this paper is structured as follows.
\Cref{ST:Sec:RelatedWork} reviews related and previous work on \ac{NST} and local-scale adjustable style transfer approaches. \Cref{ST:Sec:Method} describes our method, gives an overview of our system, and outlines implementation aspects. \Cref{ST:Sec:UserInterface} briefly explains the structure and capabilities of the user interface provided by our mobile app. \Cref{ST:Sec:Results} shows and discusses exemplary results and application examples. Finally, \Cref{ST:Sec:Conclusions} concludes this paper and provides a prospect on future work.

\section{Related Work}
\label{ST:Sec:RelatedWork}

\noindent \ac{NST} was introduced in the seminal work of Gatys \etal\cite{Gatys2016a}, which is based on an iterative optimization of feature statistics of a content, style and target image, extracted using a \ac{CNN}. 
The \enquote{similarity} of the target image to the style is represented as the difference between Gram matrices in style and target image over a set of feature responses of the VGG~\cite{simonyan2015very} network. 
However, the optimization process is computationally intensive and slow, and thus not suited for use in interactive and mobile environments. 
To approach this issue, several methods of accelerating style transfer have been published---denoted as fast style transfer methods in literature \cite{jing2019neural}. 
Generally, these methods train a network to directly inference the stylized output, such as the popular architecture introduced by Johnson \etal\cite{JohnsonAF16}.

Furthermore, feed-forward network architectures can be categorized by their ability to represent different styles. 
While the approach of Johnson \etal\cite{JohnsonAF16} can represent a single style per network, follow-up works added the ability to represent multiple styles (Zhang \etal\cite{Zhang2019}, Msg Net) or arbitrary styles (Huang \etal \ac{adaIN} \cite{Huang2017}). 
However, representing more styles with one single network generally represents a trade-off in quality, memory and run-time performance versus single-style networks \cite{Jing2018,ReimannKPSTD18}, and thus the single-style-per-network approaches have prevailed in mobile applications as the "gold standard" (\eg refer to Prisma \cite{prisma} and Becasso \cite{PasewaldtSDS16}).
While the overall goal of most style transfer methods has been to achieve plausible global results without requiring user interaction, several methods allow to adjust perceptual factors of the output in varying degrees, and thus directly or indirectly control semiotic aspects known from artwork production \cite{SID17}.
Gatys \etal\cite{GatysEBHS17} demonstrate that the iterative approach can adjust the style content trade-off and colorization through weighting different loss terms. Wu \etal\cite{wu2019direction} control stroke orientation by adding an additional direction-aware loss term to the optimization.
In the feed-forward network approaches, the controllable factors are either an inherent property of the network or have to be explicitly built into the network architecture.
For example, arbitrary \ac{NST} methods such as \ac{adaIN}~\cite{Huang2017} inherently allow adjusting the stroke size by re-scaling the input style image, while this is not possible for single-style transfer methods. 

For explicit control, Reimann \etal\cite{ReimannKPSTD18} extend a multiple-styles-per-network approach with a consistency loss to enable seamless local combination of different styles.
Jing \etal\cite{Jing2018} introduce an approach for globally and locally adjusting stroke sizes using a multi-branch network architecture that includes a Stroke-Pyramid and explicitly training different style sizes, and Yao \etal incorporate self-attention into multi-stroke transfer~\cite{yao2019attention}. 
Similarly, our approach also extends a feed-forward, single-style-per-network architecture with run-time controls.
However, in contrast to previous approaches, it allows a wider range of control for editing and is more efficient with respect to the run-time and memory consumption for large images and large-scale texture marks. 
We demonstrate how to incorporate our approach and other state-of-the-art \acp{NST} into our pipeline for interactive editing with local guidance and high-resolution upsampling in our \StyleTune app.

\begin{figure*}[tb]
\centering
\includegraphics[width=1.0\linewidth]{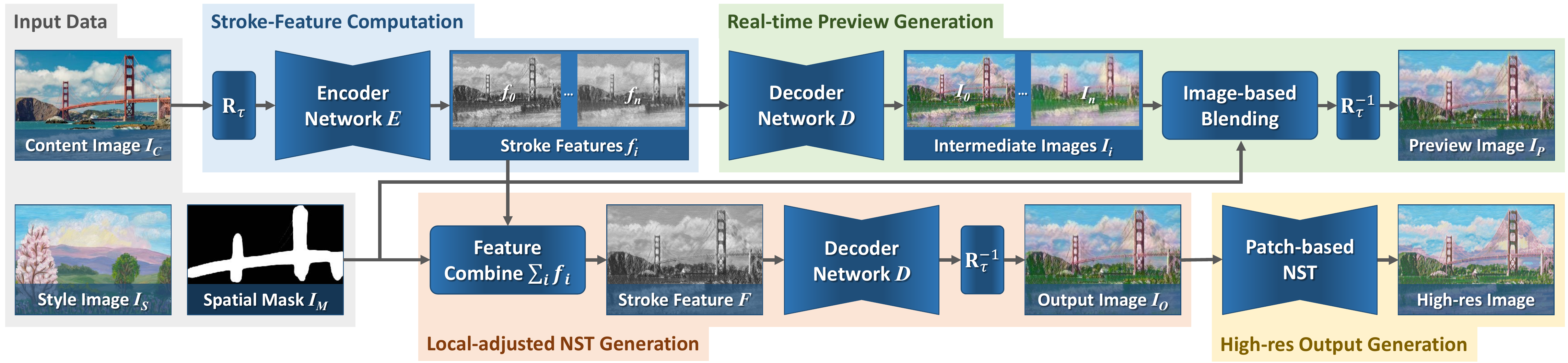}
\caption{Schematic overview of the processing stages, components, and data flow for the presented approach.}
\label{ST:Fig:Method}
\belowfig 
\end{figure*}

\section{Method}
\label{ST:Sec:Method}

\subsection{Preliminary Analysis}
\label{ST:SubSec:Preanalysis}

\noindent Fundamentally, in \ac{NST}, textural elements of the style such as stroke placement, size, or orientation are not defined explicitly, but implicitly learned by matching the style images Gram-based statistics over extracted features using image recognition networks such as VGG \cite{simonyan2015very}. These stroke textons \cite{zhu2005textons} are micro-structures in the generated image and reflect perceptual style patterns in images stylized by NST. They act as the implicit and entangled representation of several painterly concepts such as stroke/brush-size, -orientation or -intensity that are visible in the style image. In line with previous work \cite{Zhang2019}\cite{Jing2018}, controlling strokes is thus referred to adjusting stroke textons, where the visual representation of a stroke entirely depends on the extracted style statistics. 

Stroke sizes in the generated output are sensitive to the resolution of the style image, as the Gram-based statistics over the VGG features are themselves not scale-invariant. 
Resizing the \emph{style image} during training can thus be used to control the stroke size in the output image \cite{Jing2018}. Furthermore, the stroke size also strongly depends on the resolution of the \emph{input image}, as networks, once trained with a style image, produce stroke features on a certain scale dependent on the receptive field of the network. Thus, a smaller image forwarded through the same network will produce larger strokes in the output image as the receptive field of convolutional units in the network is relatively higher. Similarly, convolutional units are also not invariant against rotation of the input, \ie forwarding the rotated image through an object detection network and then rotating the output back produces different results than forwarding an un-rotated image. In the following, we show how we make use of these inherent properties of convolutional neural networks to control multivariate aspects of the style.

\beforesubsec
\subsection{Method Overview}
\label{ST:SubSec:Overview}

\noindent\Cref{ST:Fig:Method} shows an overview of our method for a controllable \ac{NST}, which allows control of stroke size, orientation, and intensity on a global and local level. It comprises the following main stages:

\newcommand{\descriptionitem}[1]{\item[]\hspace{-0.9cm} \textbf{#1:}} 
\begin{itemize}
\descriptionitem{Stroke-Feature Computation} Computes stroke features $f_i$ for different stroke granularities based on the rotated (angle $\tau$) content input image $I_C$ using the content encoder network $E$ trained on the style image $I_S$  (\Cref{ST:SubSec:NetworkArchitecture}).
\descriptionitem{Real-time Preview Generation}
To support real-time preview of local-adjustment results $I_P$, this stage performs image-based blending of the results of decoder network $D$ (\Cref{ST:SubSec:NetworkArchitecture}) applied to all features $f_i$ using the spatial mask image $I_M$ and inverts the rotation subsequently.
\descriptionitem{Local-adjusted \ac{NST} Generation} To synthesize a high-quality output image $I_O$, this stage performs model-space blending by combining the features $f_i$ based on $I_M$ into a unified stroke feature map $F$ that is fed into the decoder $D$ and followed-up with a rotation inversion (\Cref{ST:SubSec:Orientation}).
\descriptionitem{High-resolution Output Generation} To obtain a high-resolution output, this stage performs optional upsampling of $I_O$ using a service-based implementation of patch-based upsampling~\cite{Texler2019}.
\end{itemize}




\subsection{Parameter Mapping}
\label{ST:SubSec:ParameterMapping}
\noindent  Our approach combines the controllability of the stroke size and style intensity parameter (\Cref{ST:Fig:Stroke_Intensity_Grid}) by mapping them to different input modalities of the proposed network. Stroke-orientation control on the other hand can be achieved through transformation of the input image and does not need any architecture considerations.

\subsubsection{Stroke-Size Control}
 We make use of the scale dependency of receptive fields (\Cref{ST:SubSec:Preanalysis}) in our architecture to scale stroke sizes (\Cref{ST:Fig:Stroke_Intensity_Grid}). The naive approach of downscaling the image, applying a conventional fast style transfer network, and upscaling the image using a super resolution upscaling approach can achieve control over stroke sizes, as Jing \etal\cite{Jing2018} show. However, the output loses sharpness and details. Therefore, the key idea of our network is to combine dynamic input scaling with extracted high-frequency details from the high-resolution input image. The stroke size parameter $\lambda_S$ effectively controls the receptive field size of the stroke branch (\Cref{ST:SubSec:NetworkArchitecture}).
 
\subsubsection{Stroke-Intensity Control}
\noindent Based on  instance normalization, Dumoulin \etal\cite{Dumoulin45832} proposed a \ac{CIN} layer that, instead of learning affine parameters, learns a set of parameters $\beta$ and $\gamma$ for each style. In their approach, conditioning on a style is achieved as follows:
\begin{equation*}
 z = \gamma_{s}\left(\frac{x - \mu(x)}{\sigma(x)}\right) + \beta_{s}
\end{equation*}
where $\mu(x)$ and $\sigma(x)$ refer to the mean and standard deviation taken across the spatial dimensions of input features $x$. The \ac{NST} network incorporates multiple \ac{CIN} layers, and sets $\beta_{s}$ and $\gamma_{s}$ according to the chosen style $s$.

\begin{figure}[tb]
\tikzset{every node/.style={inner sep=0pt,outer sep=0pt}}
\settoheight{\tempdima}{\includegraphics[width=\halfFigSize]{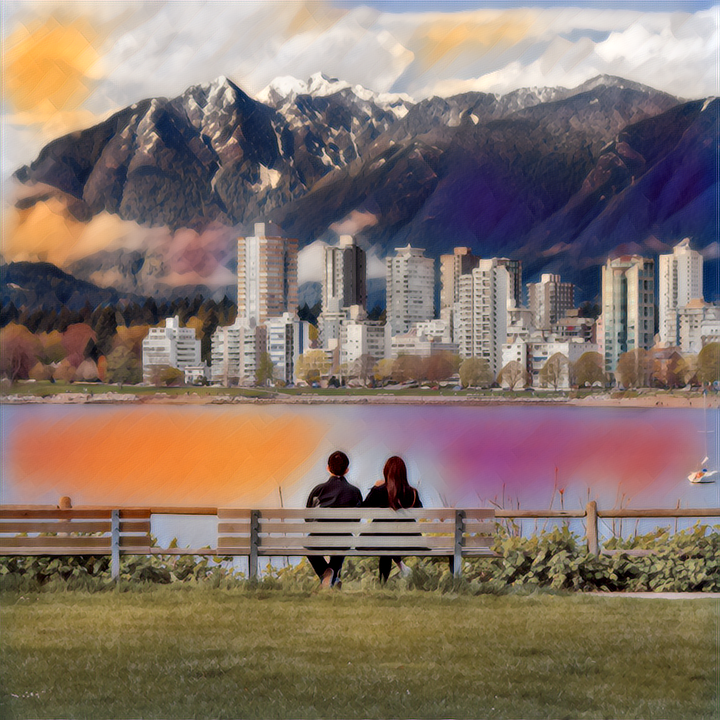}}%
	\centering\begin{tabular}{cc@{}c@{}c@{}c@{}}
		& {$\lambda_I=0.2$} & {$\lambda_I=1.0$} & {$\lambda_I=3.0$} \\
		\rowname{$\lambda_S=1.0$}&
		\begin{tikzpicture}
		\node{\includegraphics[width=\halfFigSize]{resources/grid_scale_intens/picabia_standard_1.0intens_0.2.png}};%
		\end{tikzpicture}&
		\begin{tikzpicture}
		\node{\includegraphics[clip,width=\halfFigSize]{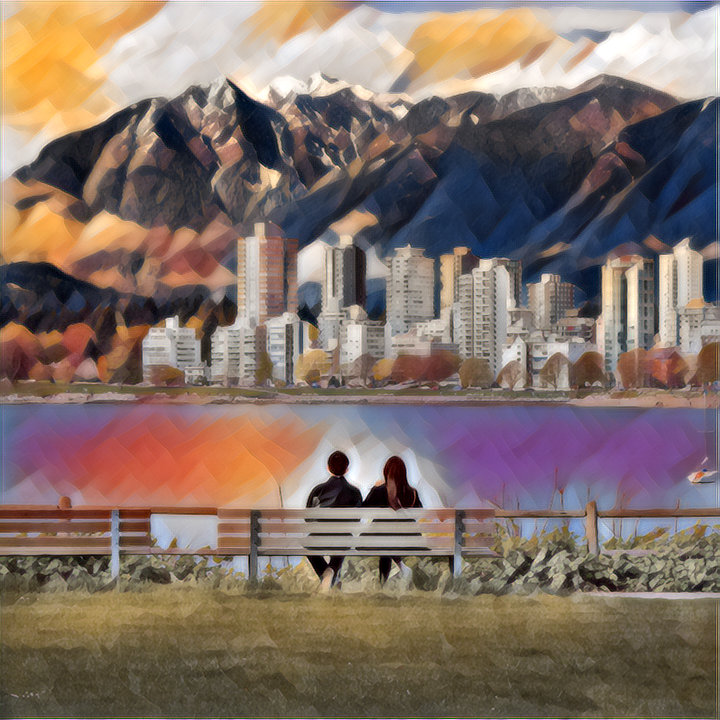}};%
		\end{tikzpicture}&
		\begin{tikzpicture}
		\node{\includegraphics[width=\halfFigSize]{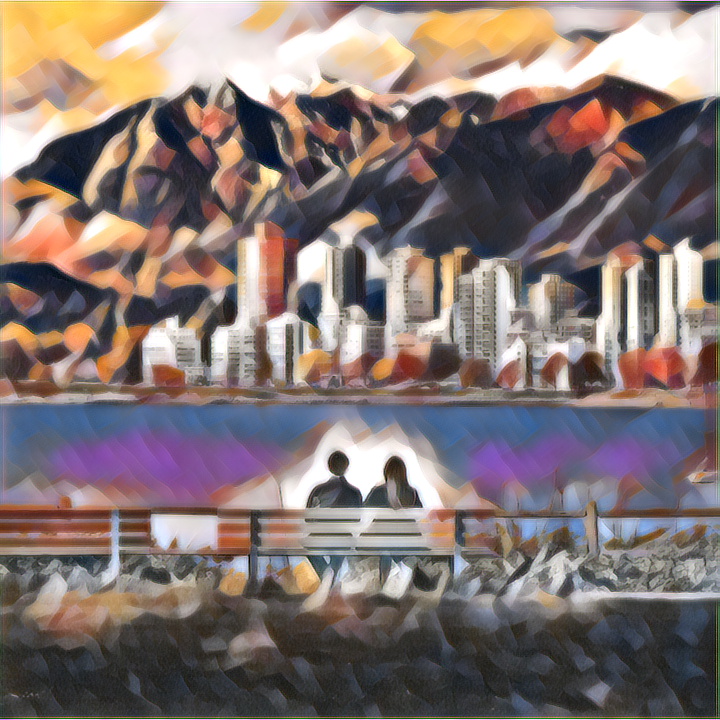}};%
		\end{tikzpicture}\\[-1.2ex]
		\rowname{$\lambda_S=2.0$}&
		\begin{tikzpicture}
		\node{\includegraphics[width=\halfFigSize]{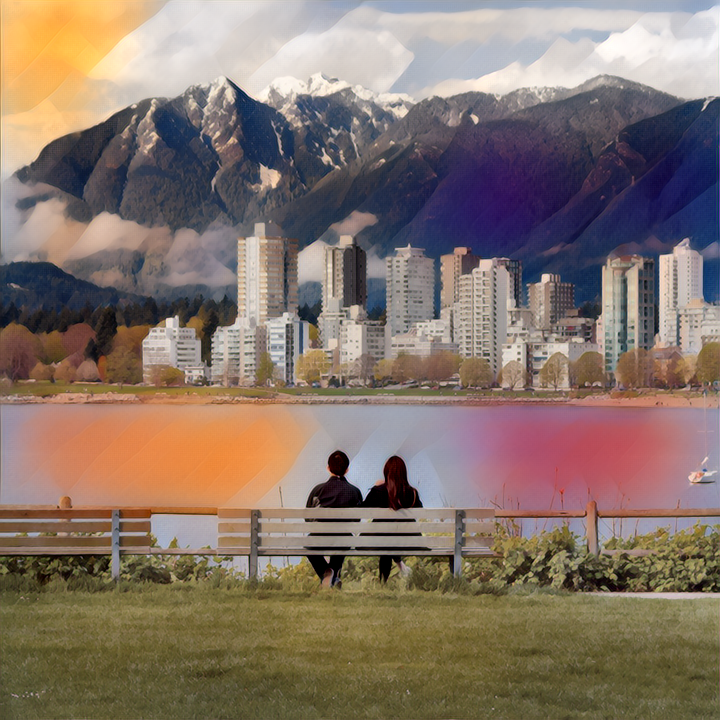}};%
		\end{tikzpicture}&
		\begin{tikzpicture}
		\node{\includegraphics[width=\halfFigSize]{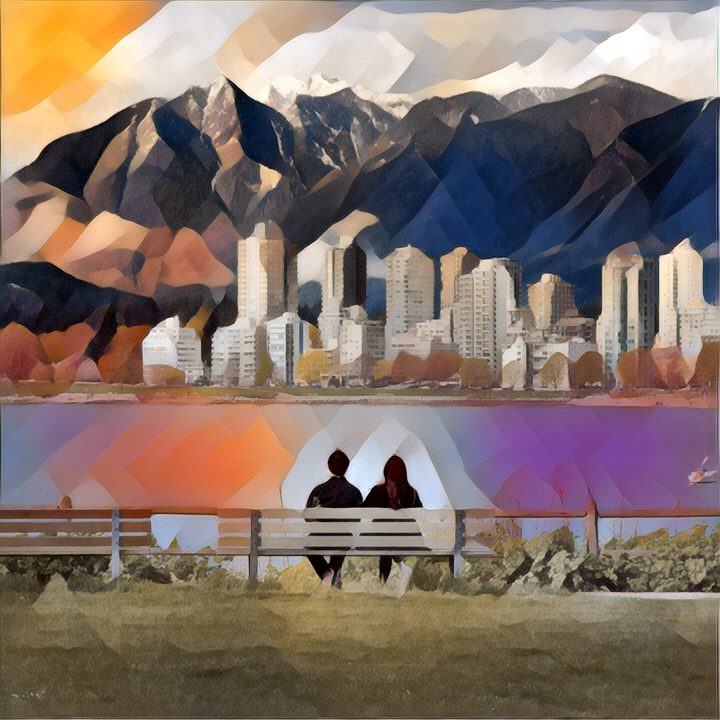}};%
		\end{tikzpicture}&
		\begin{tikzpicture}
		\node{\includegraphics[width=\halfFigSize]{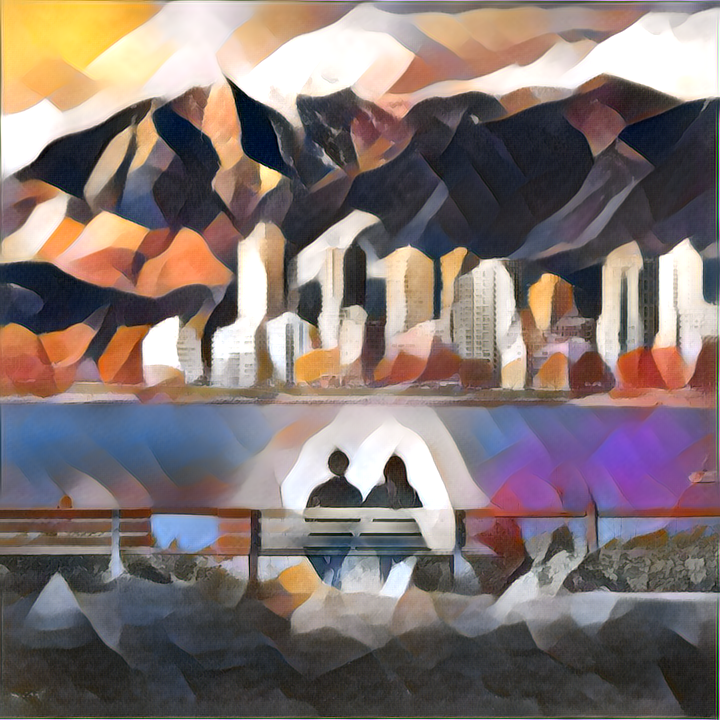}};%
		\end{tikzpicture}\\[-1.2ex]
		\rowname{$\lambda_S=3.0$}&
		\begin{tikzpicture}
		\node{\includegraphics[width=\halfFigSize]{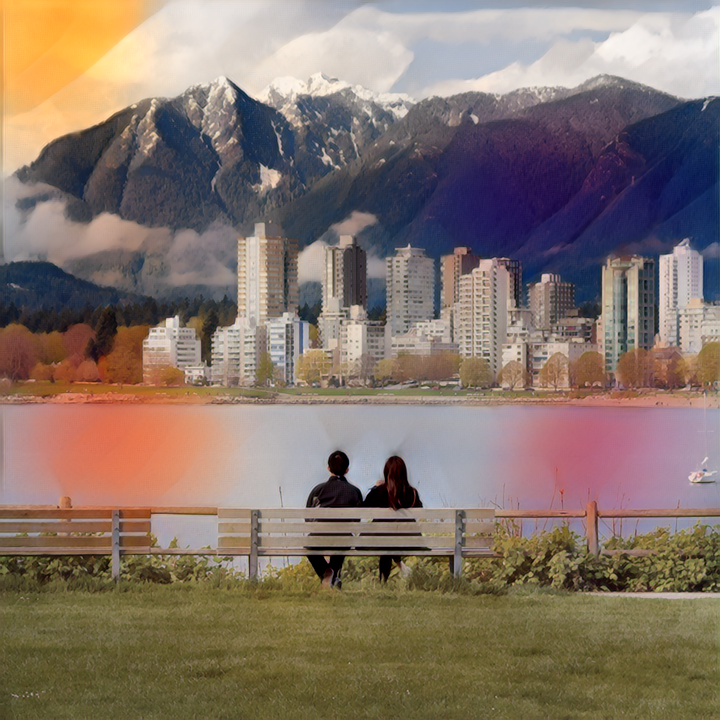}};%
		\end{tikzpicture}&
		\begin{tikzpicture}
		\node{\includegraphics[width=\halfFigSize]{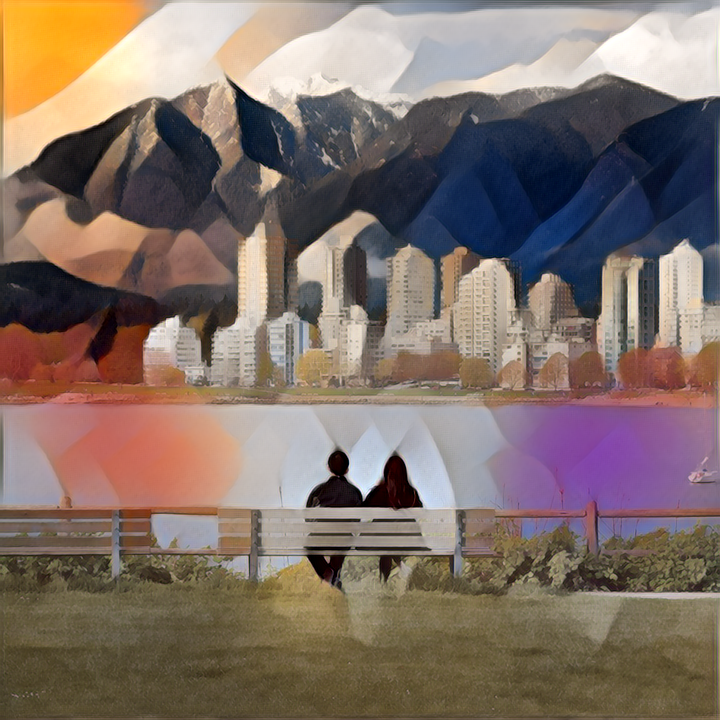}};%
		\end{tikzpicture}&
		\begin{tikzpicture}
		\node{\includegraphics[width=\halfFigSize]{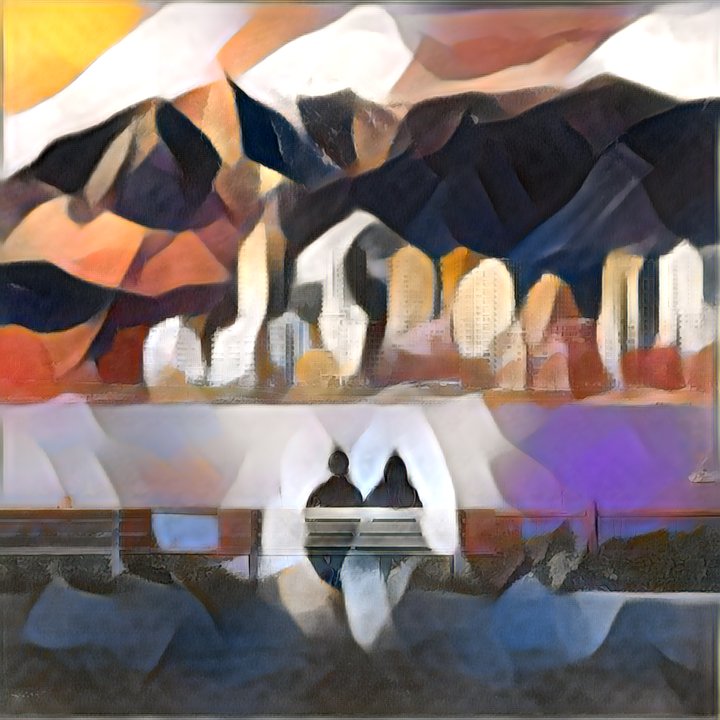}};%
		\end{tikzpicture}
	\end{tabular}
	\caption{Variations of stroke size $\lambda_S$ and style intensity $\lambda_I$.}
	\label{ST:Fig:Stroke_Intensity_Grid}
	\belowfig 
\end{figure}

Huang \etal\cite{Huang2017} further show, that the parameters can be directly predicted from an input style image, without requiring to train the corresponding style, yielding an \ac{adaIN} layer. We adapt the parametrization of $\alpha$ and $\beta$ as a form of general controllability that can map continuous input variables to desired network outputs. Similar to Babaeizadeh \etal\cite{babaeizadeh2019adjustable}, we map style intensity (\Cref{ST:Fig:Stroke_Intensity_Grid}) via the \ac{CIN} layers, where the set of all \ac{CIN} parameters $\Phi = \{\alpha, \beta\}$ is predicted using a linear regression as follows:
 \begin{equation*}
 \Phi = W \lambda_I + b
 \end{equation*}
 where $\lambda_I$ is the input style intensity and $W, b$ are learnable weights obtained using a fully-connected layer.
  
\setlength{\tabcolsep}{0.5mm} 

\begin{figure}[tb]
\centering
\includegraphics[width=1.0\linewidth]{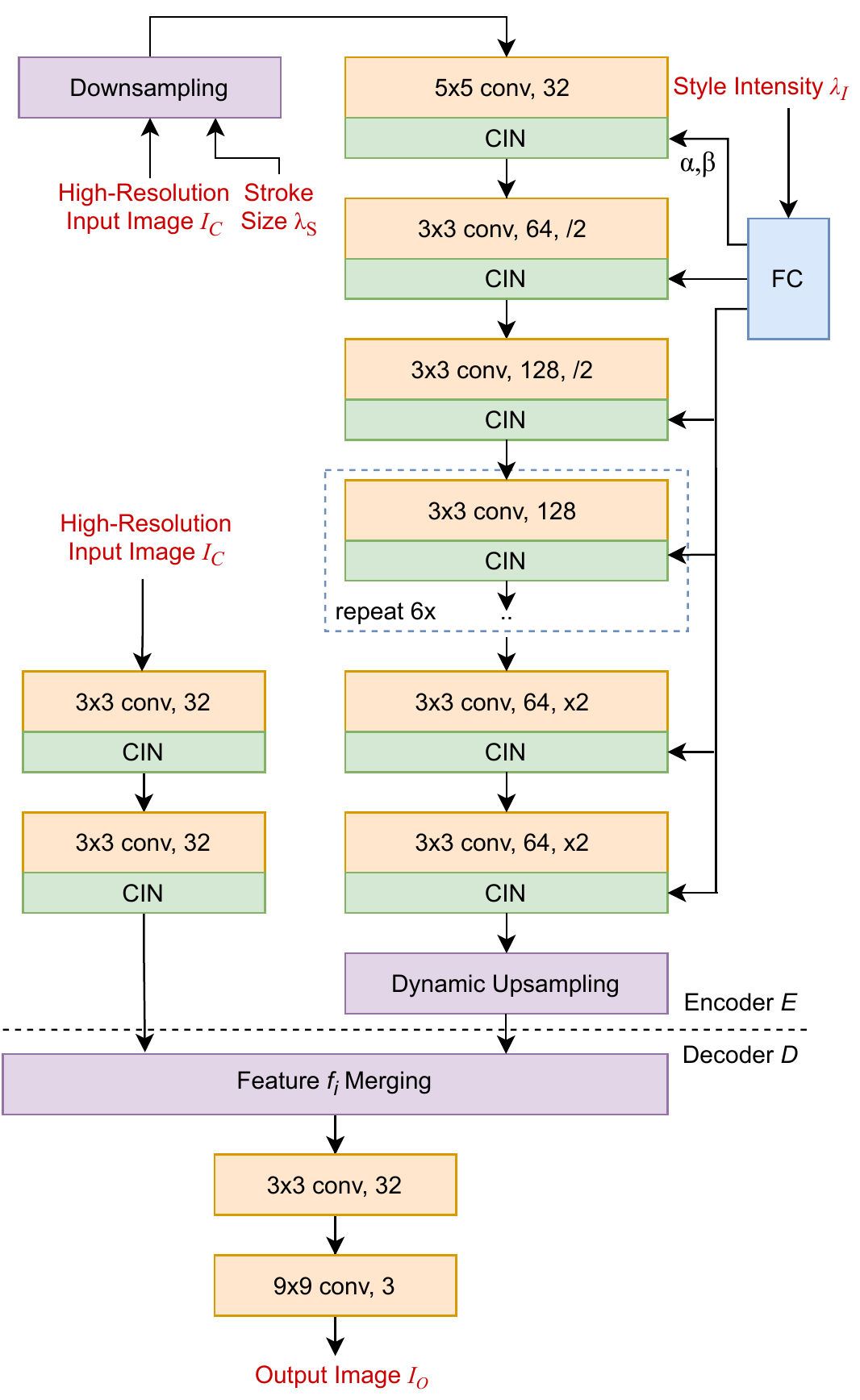}
\caption{Adjustable \ac{NST} network architecture. The encoder network $E$ consists of a high-resolution processing branch (left) and a low-resolution branch (right), that produce features $f_i$ for a certain stroke size. The decoder $D$ consists of a feature merging and decoding operation. The stroke scale $\lambda_S$ is used as the downsampling factor of the input image for the low-resolution branch. For each residual block (depicted in orange), kernel size, channel number, and spatial up/down sampling are shown. Residual blocks use \ac{CIN} operations with parameters $\alpha, \beta$ controlled through a regression from the style-intensity parameter $\lambda_I$.}
\label{ST:Fig:Architecture}
\belowfig 
\end{figure}

\subsection{Network Architecture}
\label{ST:SubSec:NetworkArchitecture}

\noindent \Cref{ST:Fig:Architecture} shows our proposed adjustable \ac{NST} network architecture. 
The network consists of a two-branch architecture using a Resnet-like~\cite{he2016deep} structure with residual-blocks. The dynamic/low-resolution branch is based on the fast style transfer architecture~\cite{JohnsonAF16}, with instance normalization replaced by \ac{CIN} layers, and is responsible for learning the stylization operation. The high-resolution branch consists of a relatively light-weight set of layers for extracting high-frequency details from the input image. Both branches are fused together by channel concatenation before decoding the final output image. The input stroke size $\lambda_S$ controls the downsampling factor on the input image, with the dynamic upsampling operation used to upsample back to the original resolution before merging features. For local editing, the encoder part $E$ of the network is used to produce features $f_i$ in different style scales which are blended together based on a spatial mask $I_M$ during feature merging~(\cf \Cref{ST:Fig:Method}) and then jointly decoded by the network decoder~$D$. 

\subsection{Network Training}
\label{ST:SubSec:NetworkTraining}

\noindent We train our proposed network on the MS-COCO dataset~\cite{Lin:2014:MCC} for \num{4} epochs using the Adam optimizer \cite{KingmaB14}. Images are cropped and resized to $512 \times 512$ pixels, and the pre-trained VGG-\num{19} \cite{simonyan2015very} network is used as the loss network, using style and content loss as defined by Gatys \etal\cite{Gatys2016a}, for which ${relu1\_1, relu2\_1, relu3\_1, relu4\_1, relu5\_1}$ are used as the style layers and $relu4\_2$ is used as the content layer. During training, the input image for the low resolution branch is cycled through different factors for downsampling. We set the downsampling factors to \num{2} and \num{4} during training. During inference time, the downsampling factor can be chosen from a continuous range even if these input sizes were not observed during training. The style intensity parameter is randomly sampled from a uniform distribution $\lambda_I \in U(0,1)$ with the corresponding style weight set to the same value.

\begin{figure}[tb]
\centerline{
\subfloat[Content/Style]{%
\includegraphics[width=0.325\linewidth]{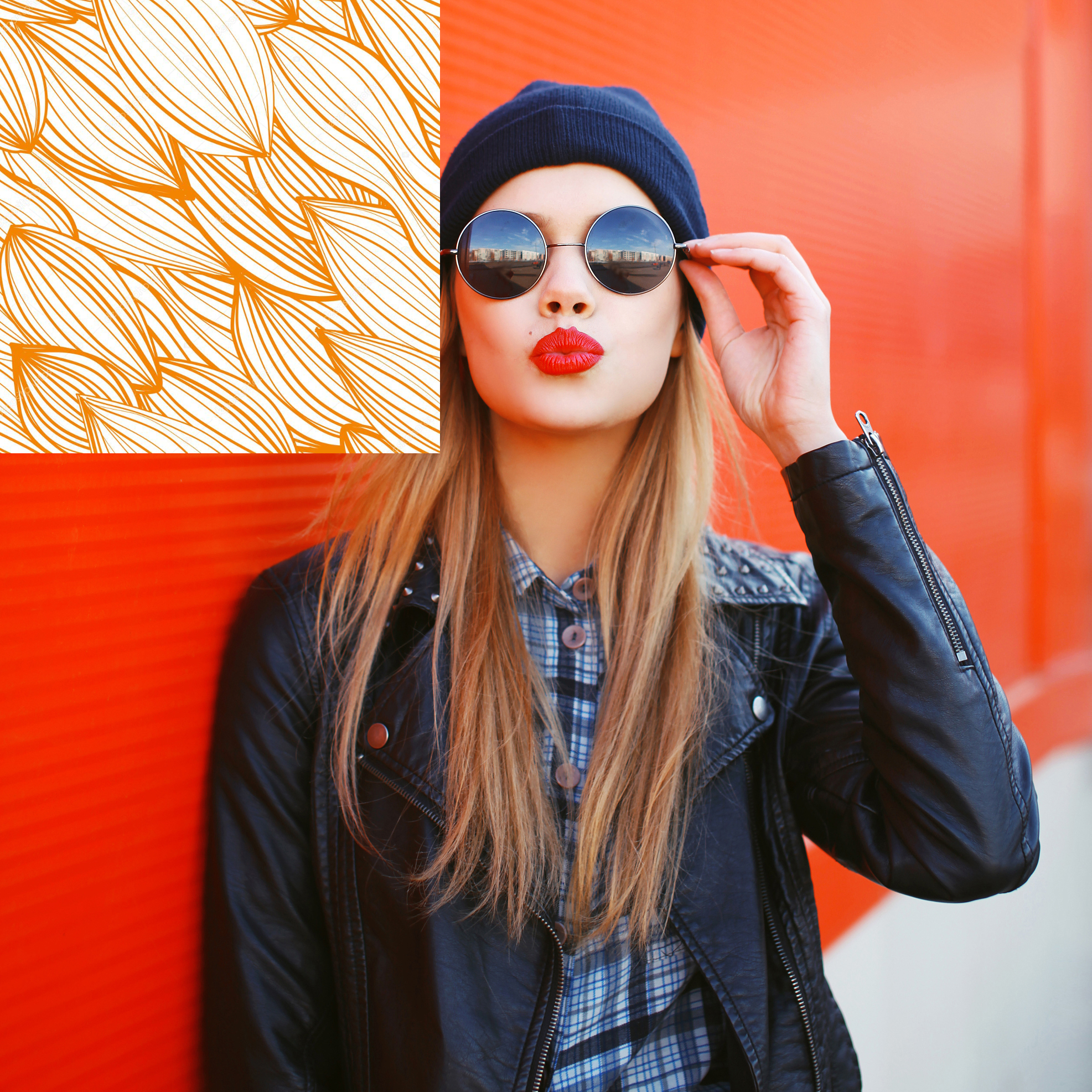}%
\label{ST:Fig:RotContentStyle}%
}\hfill
\subfloat[$\tau=0^{\circ}$ rotation]{%
\includegraphics[width=0.325\linewidth]{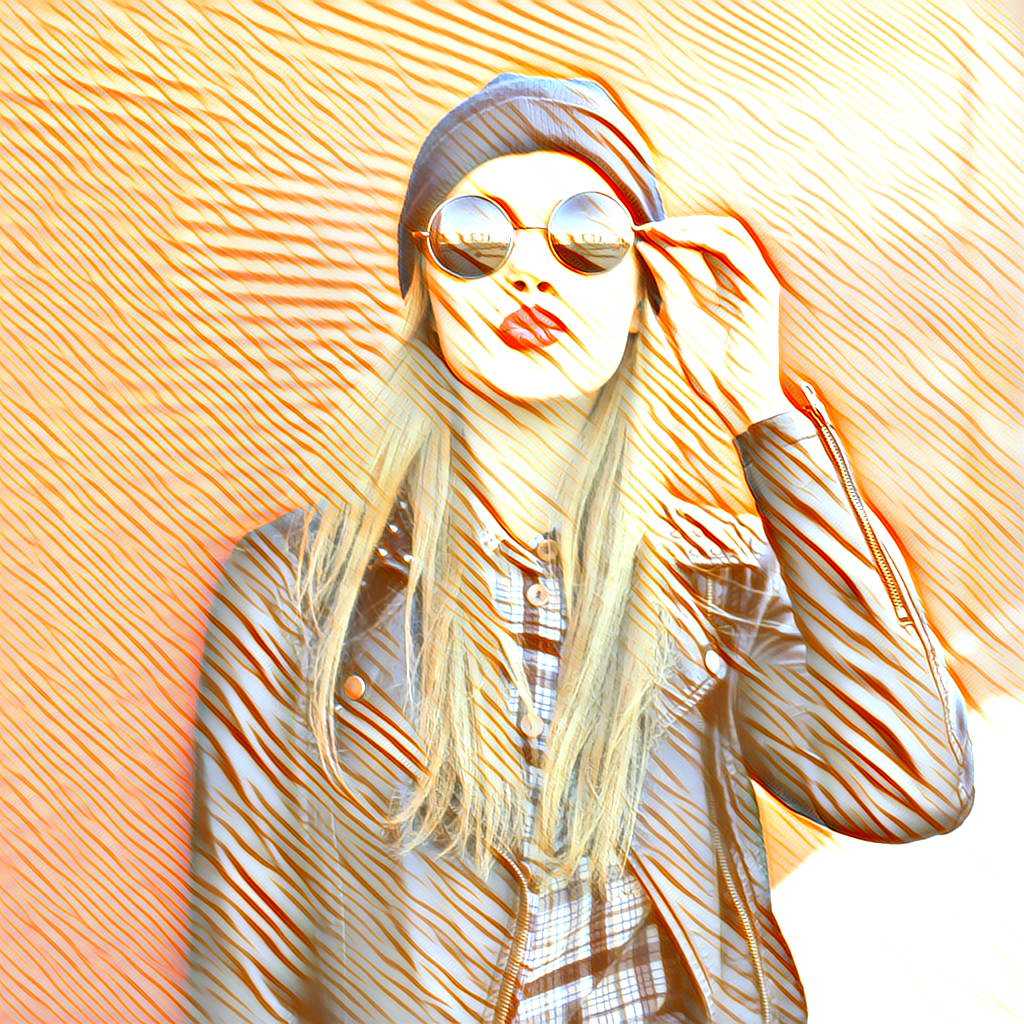}%
\label{ST:Fig:Rot1}%
}\hfill
\subfloat[$\tau =45^{\circ}$ rotation]{%
\includegraphics[width=0.325\linewidth]{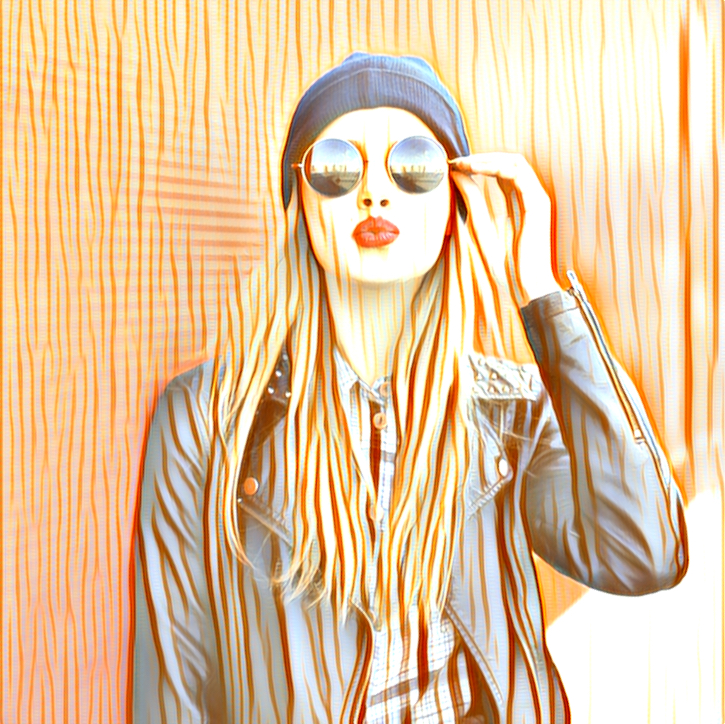}%
\label{ST:Fig:Rot2}%
}}
\caption{Change of stroke orientation by content rotation.}
\label{ST:Fig:rotating}%
\belowfig 
\end{figure}

\begin{figure*}[tp]
\centerline{
\subfloat[Content and style image selection]{%
\includegraphics[width=0.325\linewidth]{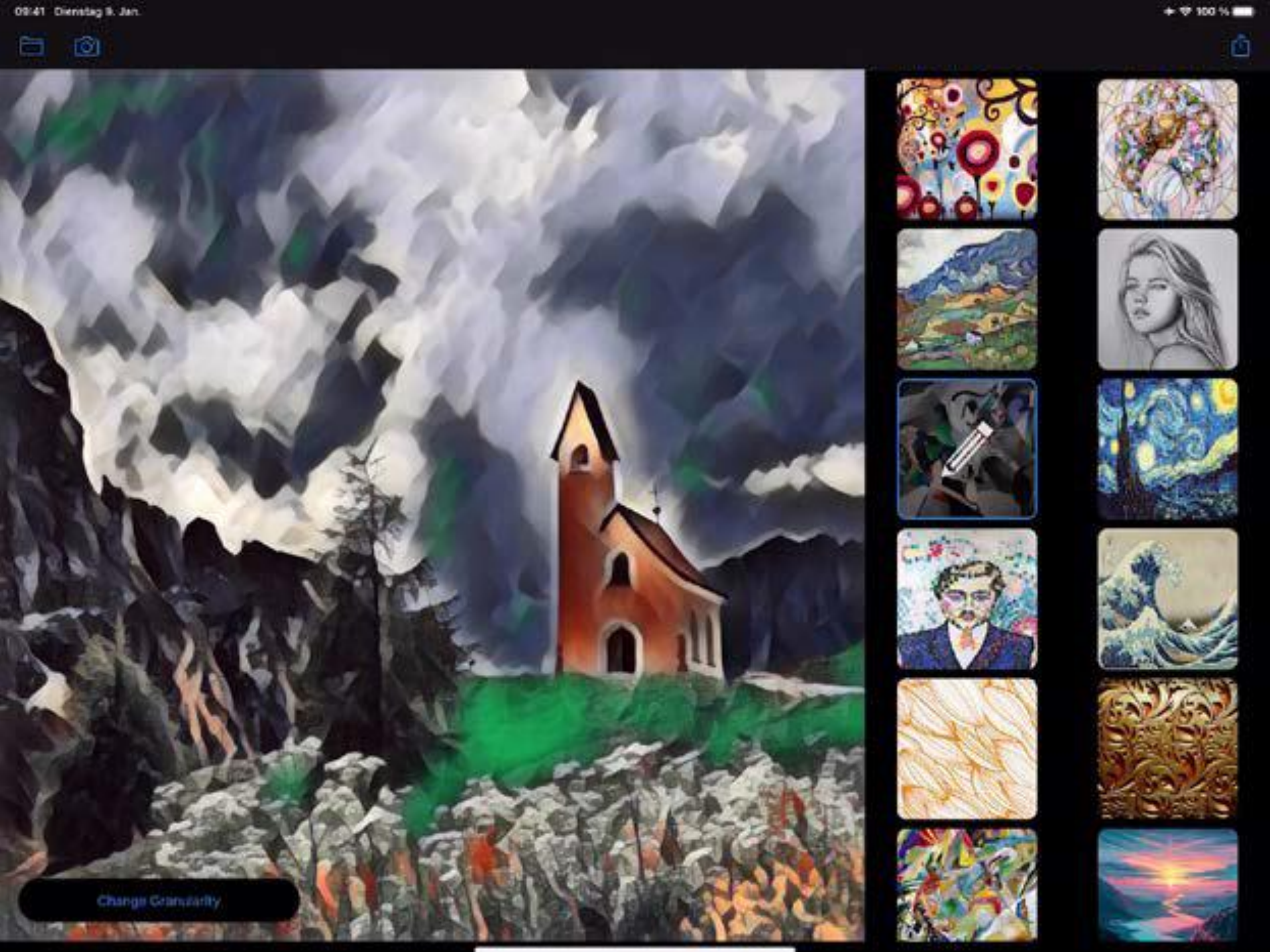}%
\label{ST:Fig:UI:A}%
}\hfill
\subfloat[Style transfer without stroke rotation.]{%
\includegraphics[width=0.325\linewidth]{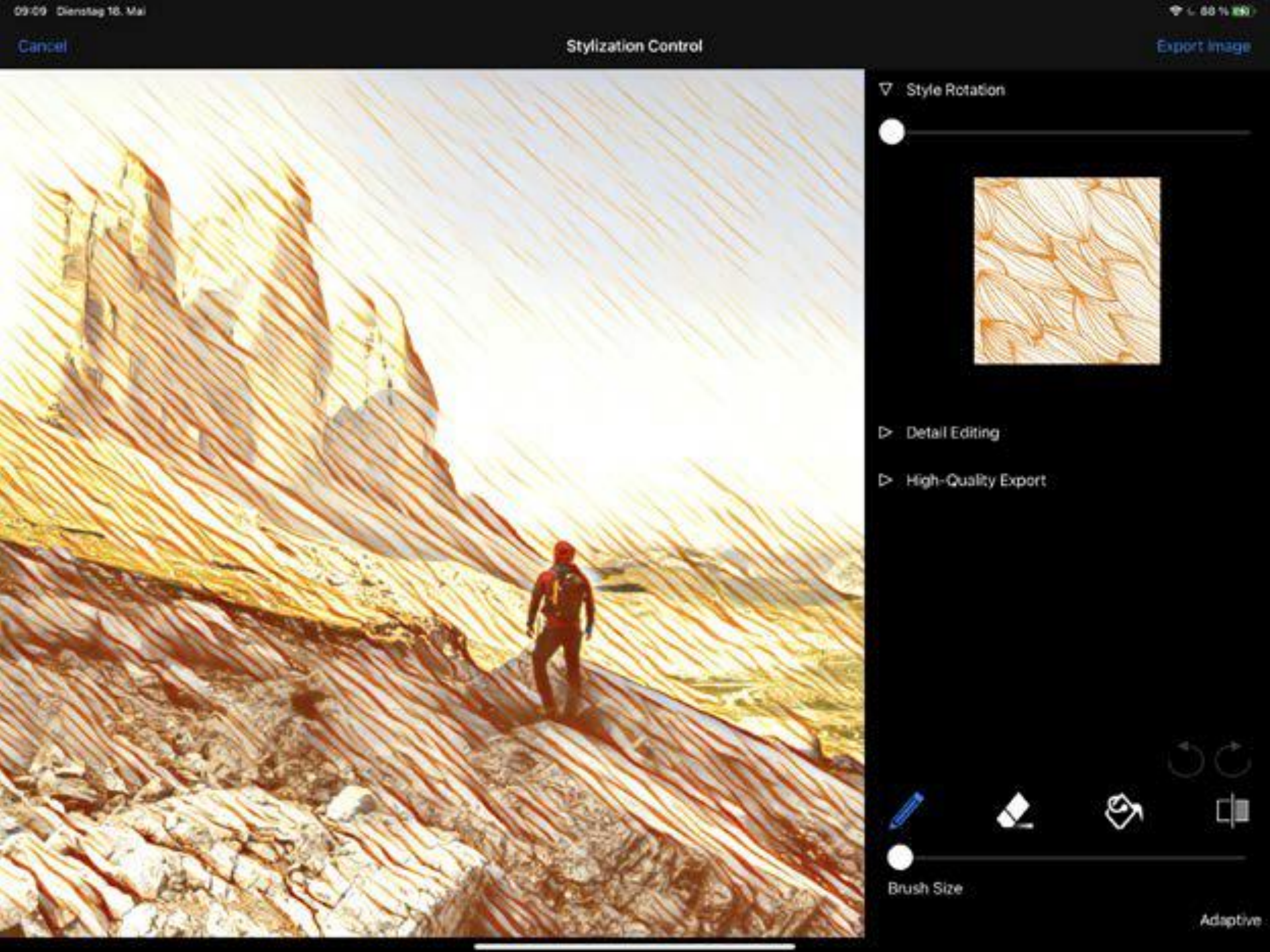}%
\label{ST:Fig:UI:B}%
}\hfill
\subfloat[Local retouches with stroke rotation.]{%
\includegraphics[width=0.325\linewidth]{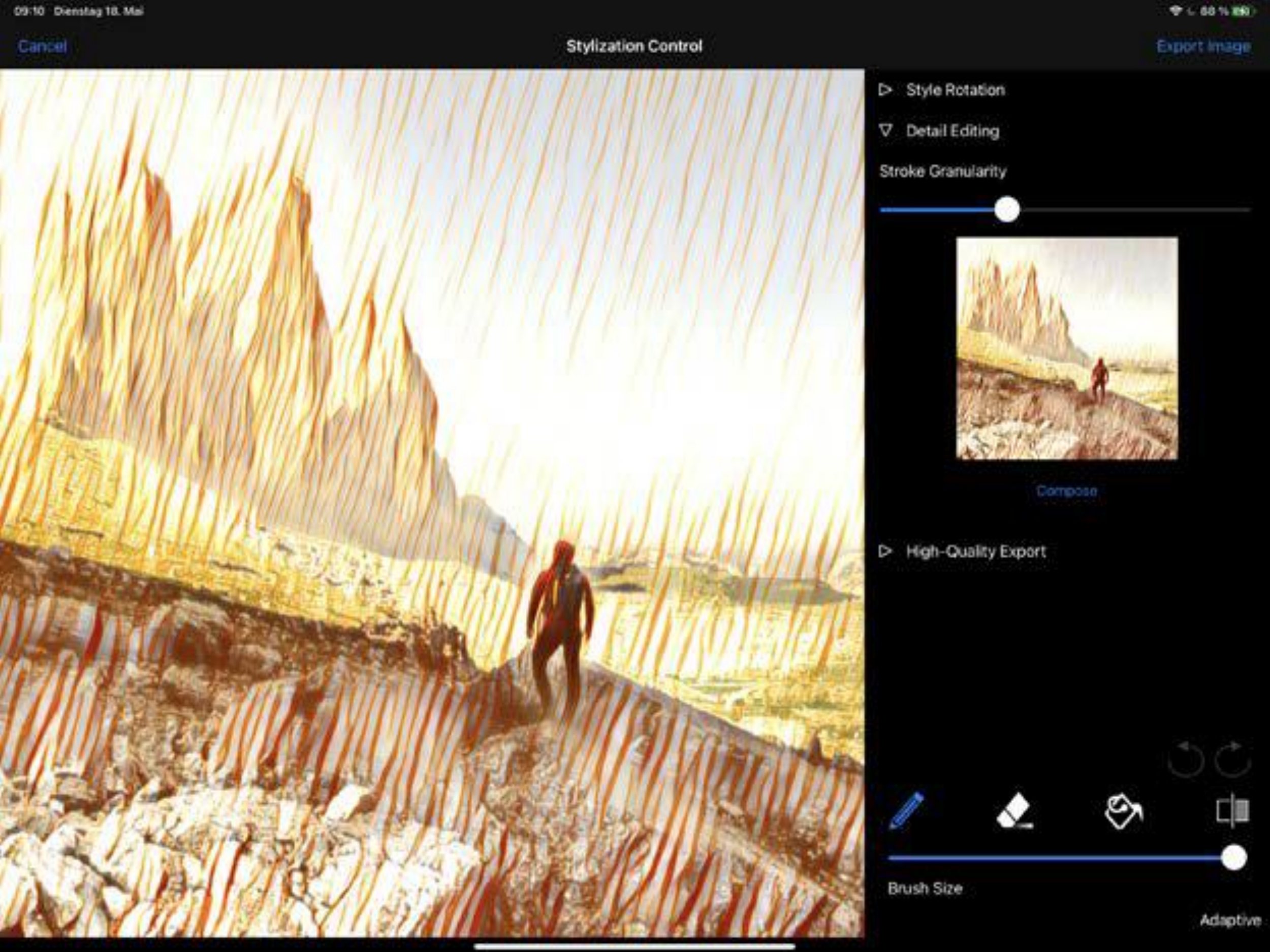}%
\label{ST:Fig:UI:C}%
}}
\caption{Screenshots of \StyleTune: After selecting a style, it can be rotated globally. The user can apply different style sizes to different parts of the image using {\itshape brush metaphors} (all content images from \textcopyright{} Shutterstock, used with permission).}
\label{ST:Fig:UserInterface}
\end{figure*}

\subsection{Stroke Orientation}
\label{ST:SubSec:Orientation}
\noindent We control stroke orientation by taking advantage of the property that convolutional units are not invariant with respect to input rotation, as described in \Cref{ST:SubSec:Preanalysis}. In particular, using the stroke rotation factor $\tau$, a stroke in the output is rotated as follows:
\begin{align*}
f_i &= E\big[\text{pad}\left(\mathbf{R}_{\tau}I_C\right)\big] \\
I_O &= \text{crop}\left[\mathbf{R}^{-1}_{\tau}D\left(\sum_i{f_i}\right)\right]
\end{align*}
where the input image $I_C$ is rotated by $\tau$ degrees using a rotation matrix $\mathbf{R}_{\tau}$ and then padded back to a rectangular shape (\ie filling whitespace resulting from the rotation) using reflection padding. The image is then fed through the encoder $E$ to obtain stroke features $f_i$, which can then be blended according to a spatial mask and fed to the decoder~$D$ to obtain the output image. The rotation is then inverted and the resulting image is cropped to the original extent to obtain a stylized image where the strokes are rotated by~$\tau$~(\Cref{ST:Fig:rotating}). This method of stroke orientation adjustment is agnostic of network architecture and can be applied using any feed-forward \ac{NST} method. 

\beforesubsec
\subsection{Implementation Aspects}
\label{ST:Sec:ImplementationAspects}

\noindent Training and model preparation is implemented in python and using PyTorch \cite{PyTorch2019}. The respective models 
are converted to CoreML~\cite{Marques2020} and weights are quantized to \SI{16}{\bit} in a pre-processing step for their use on mobile devices supporting iOS. The app implementation is based on Apple’s Swift, CoreML (MTLGPUFamilyApple5), and Metal \acp{API} for \ac{GPU}-based processing of neural networks and rendering techniques, and implements the respective functionality for local adjustments. The resulting images can then be transferred to a service-based implementation of patch-based upsampling~\cite{Texler2019}.


\section{User Interface}
\label{ST:Sec:UserInterface}

\noindent \Cref{ST:Fig:UserInterface} shows screenshots of our application illustrating the three step process to create the final output image. 
Thereby, the interactive image editing and enhancement workflow comprises the following steps and interface components
to facilitate edits on different levels-of-control~\cite{Isenberg2016}.

\paragraph{1) Selection of Content and Style Images}

\noindent After selecting and loading or capturing an input content image $I_C$, the user is required to select an style image $I_S$ from a list of available/trained styles (\Cref{ST:Fig:UI:A}). Upon style selection, the application applies the respective \ac{NST} model to the content image and presents a preview $I_P$ of the resulting style transfer in real-time. 
This enables a user to quickly browse the available styles and to decide for a basic style for his subsequent edits.

\paragraph{2) Adjustment of Stroke Granularity and Rotation}
\label{ST:SubSubSec:Granularity and RotationAdjustment}

\noindent After selecting a model for the global transfer, the user can control the global stroke size $\lambda_S$ interactively using a slider. To allow for further control, we present the user an additional optional editing view with global style settings that enables the interactive adjustment of stoke size and rotation (\Cref{ST:Fig:UI:B}). To implement such interactive exploration of the parameters (\eg using a slider), results are pre-generated when entering the edit view. Pre-computation incurs a brief loading time, depending on the number of pre-computed stroke-size levels. Furthermore, using painting brush metaphors to manipulate the spatial mask $I_M$, the user can then locally apply different stroke-size and orientation edits. The stroke levels are blended in image space to retain interactivity during the brush process and then merged in feature space of the neural network to create seamless stroke transitions on demand~(\Cref{ST:Fig:UI:C}). 

\paragraph{3) High-resolution Image Export}
\label{ST:SubSubSec:HighResolutionExport}

\noindent In a final step, the composition can be exported at a very high spatial resolution using patch-based upsampling~\cite{Texler2019}. After the upsampled image is received, the user can explore the result using pan and zoom gestures. If the user is satisfied with the result, the application allows storing and sharing respectively.

\begin{figure*} [tb]
\settoheight{\tempdima}{\includegraphics[width=\specFigSize]{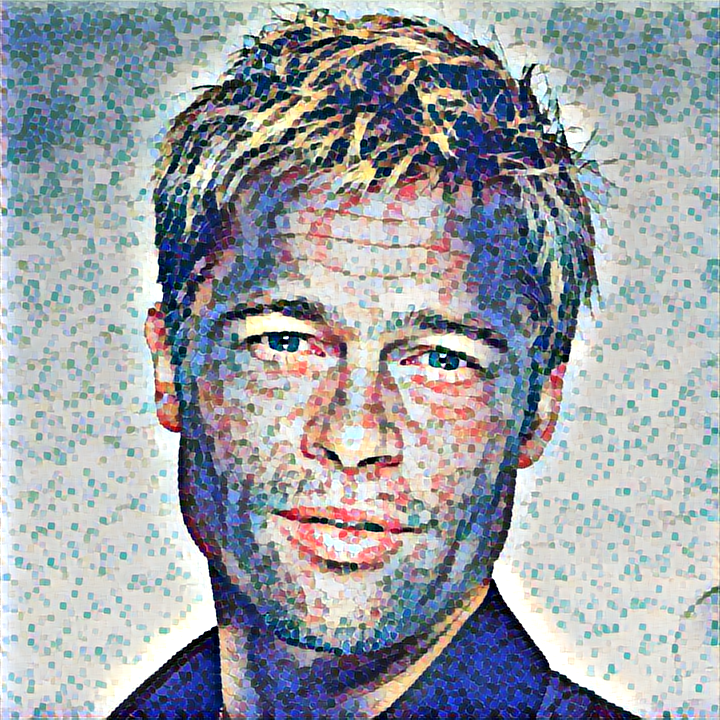}}%
\centering\begin{tabular}{@{}c@{}c@{}c@{}c@{}c@{}c@{}}
	\rowname{Jing \protect\etal \protect\cite{Jing2018}}&
	\begin{tikzpicture}
	\node{\includegraphics[width=\specFigSize]{resources/stroke-comparison/delaunay_adaptive_0.0.png}};%
	\end{tikzpicture}&
	\begin{tikzpicture}
	\node{\includegraphics[width=\specFigSize]{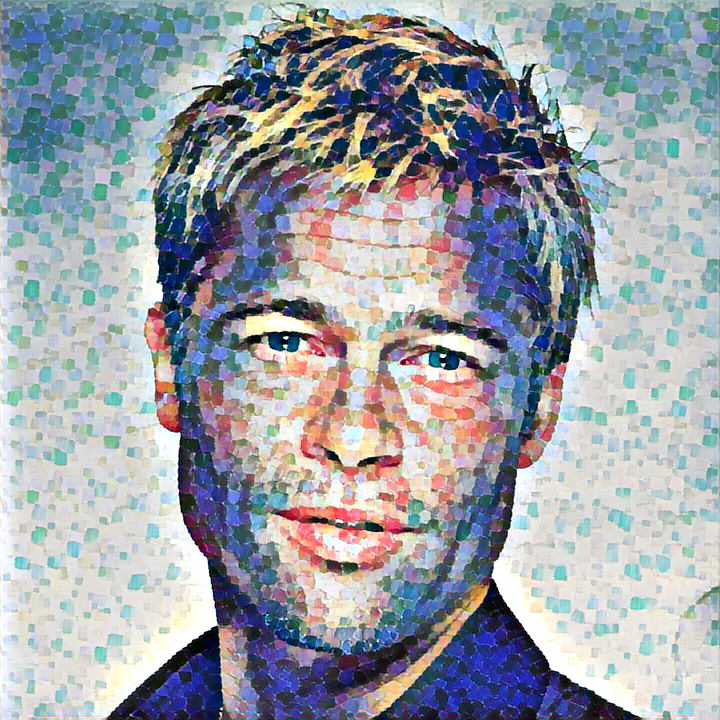}};%
	\end{tikzpicture}&
	\begin{tikzpicture}
	\node{\includegraphics[width=\specFigSize]{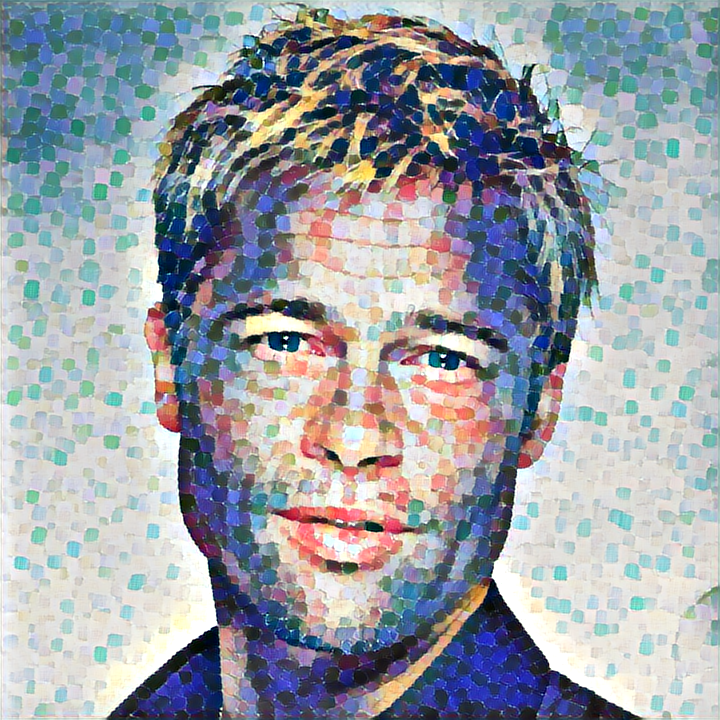}};%
	\end{tikzpicture}&
	\begin{tikzpicture}
	\node{\includegraphics[width=\specFigSize]{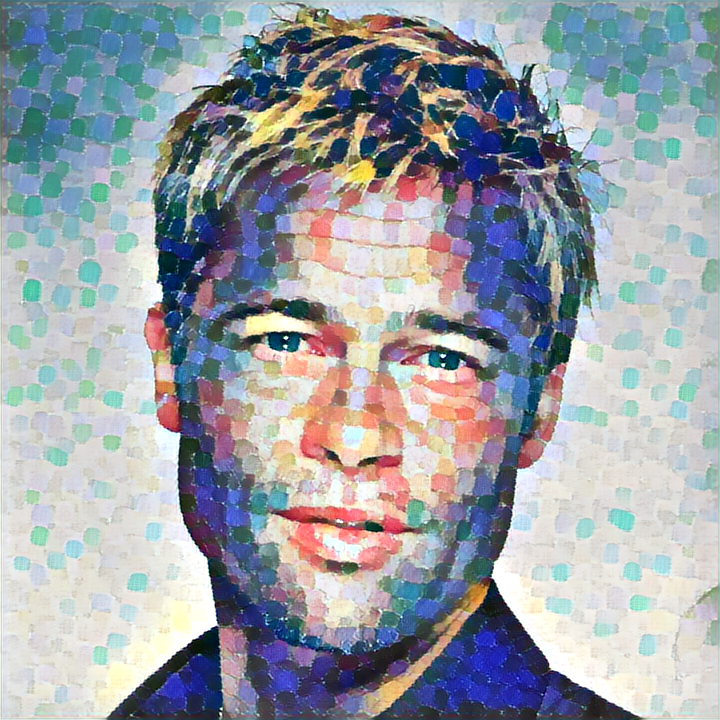}};%
	\end{tikzpicture}\\[-1.2ex]
    \rowname{Ours}&
	\begin{tikzpicture}
	\node{\includegraphics[width=\specFigSize]{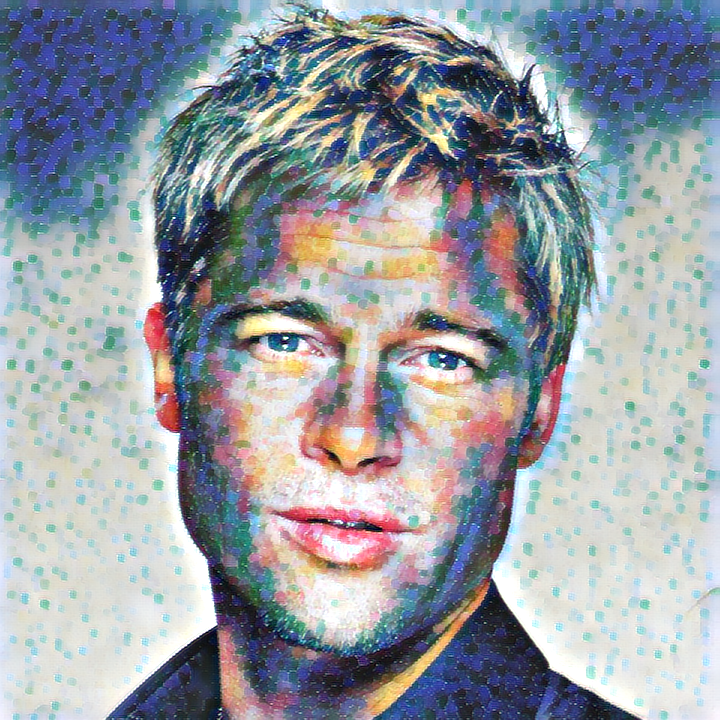}};%
	\end{tikzpicture}&
	\begin{tikzpicture}
	\node{\includegraphics[width=\specFigSize]{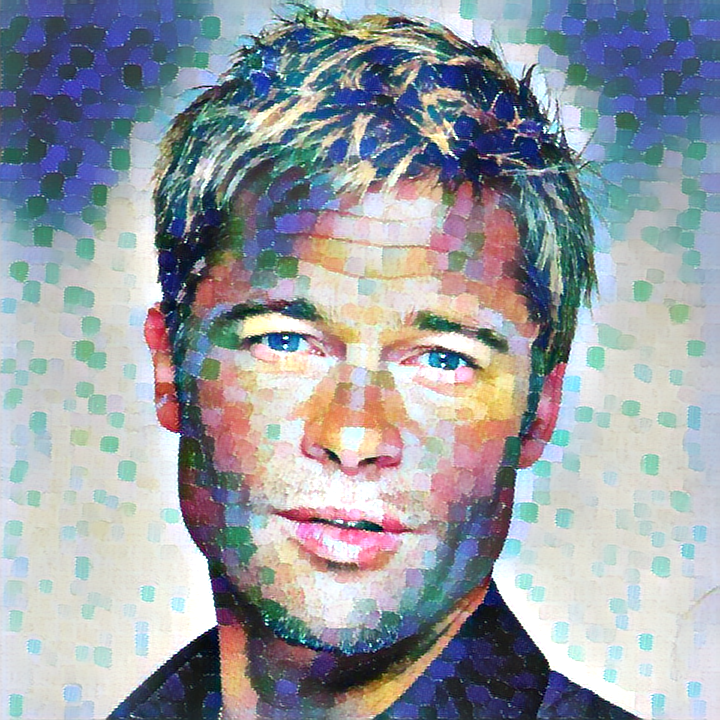}};%
	\end{tikzpicture}&
	\begin{tikzpicture}
	\node{\includegraphics[width=\specFigSize]{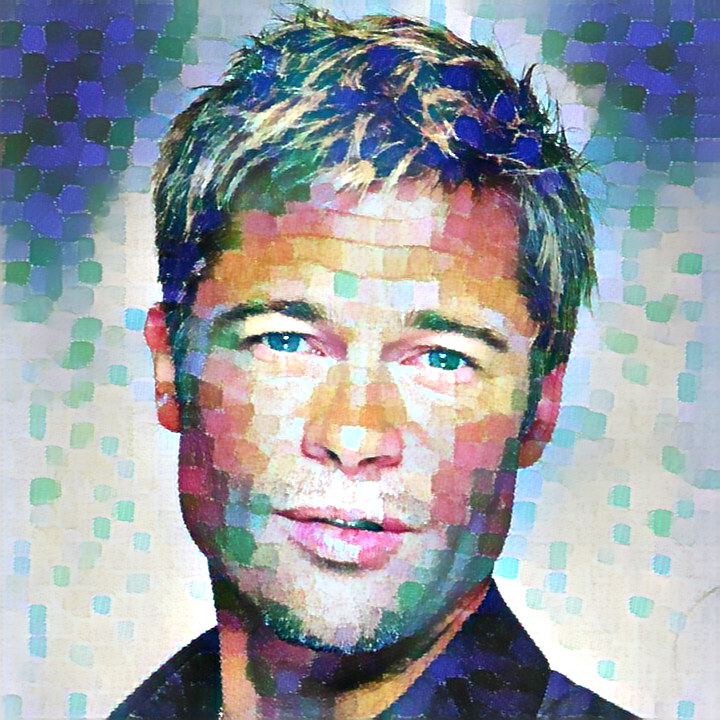}};%
	\end{tikzpicture}&
	\begin{tikzpicture}
	\node{\includegraphics[width=\specFigSize]{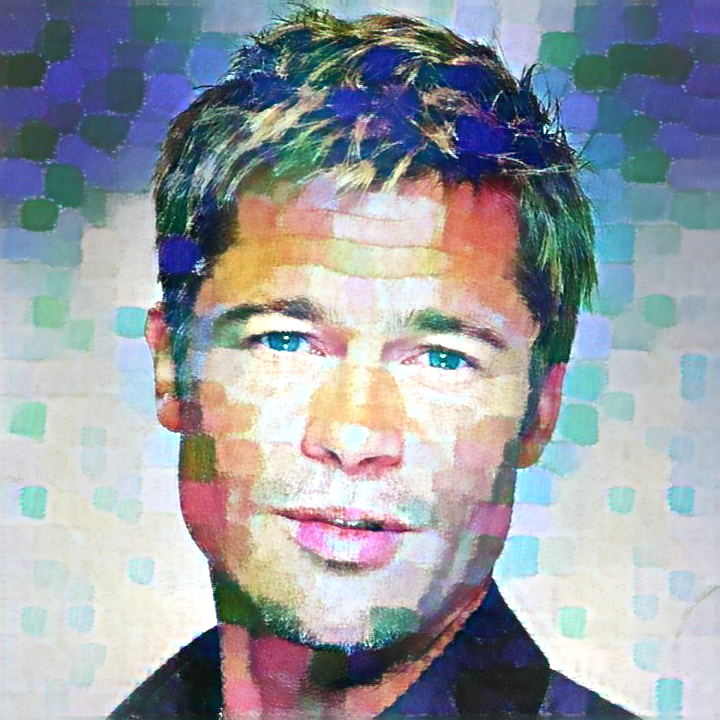}};%
	\end{tikzpicture}
\end{tabular}
\caption{Comparing global stroke size adjustment from lowest to highest level. Our approach can represent a higher range of stroke sizes than the adaptive network of Jing~\etal\cite{Jing2018}, as (in theory) arbitrarily large stroke sizes $\lambda_I$ can be set. In practice, very large stroke sizes tend to lose sharpness, as the style-branch resolution decreases.  }
\label{fig:global_size_comparison}
\belowfig 
\end{figure*}

\section{Results}
\label{ST:Sec:Results}

\subsection{Qualitative Comparison}
\label{ST:SubSection:qualcomparison}

\noindent \Cref{fig:global_size_comparison} shows comparison results of our architecture to the architecture of Jing~\etal\cite{Jing2018} as it is the most similar to our work in representing stroke size control for a single style in one feedforward style transfer network. In addition to being able to control further elements of the style such as its intensity, our adjustable architecture allows to represent a greater range of stroke size edits. This is especially true for higher resolution images, as style transfer networks are often trained on small resolutions images (usually $256 \times 256$ pixels) to reduce the computational cost, resulting in very fine and subtle style elements during inference on large images (\ie edge length \num{1024} and above). Our architecture mitigates this by downsampling the input image in the stylization branch while retaining high-frequency details in the high-resolution branch. Furthermore, our architecture is able to represent consistent stroke sizes in different output image resolutions -- if the input image resolution doubles the stroke size can be equally doubled. This can be important in an application context when working on a lower resolution preview image and exporting to a high-resolution final image.

\subsection{Exemplary Results}
\label{ST:SubSection:ExemplaryResults}

\begin{figure*}[tb]
\centerline{
\subfloat[Content and Mask]{%
\includegraphics[width=0.195\linewidth]{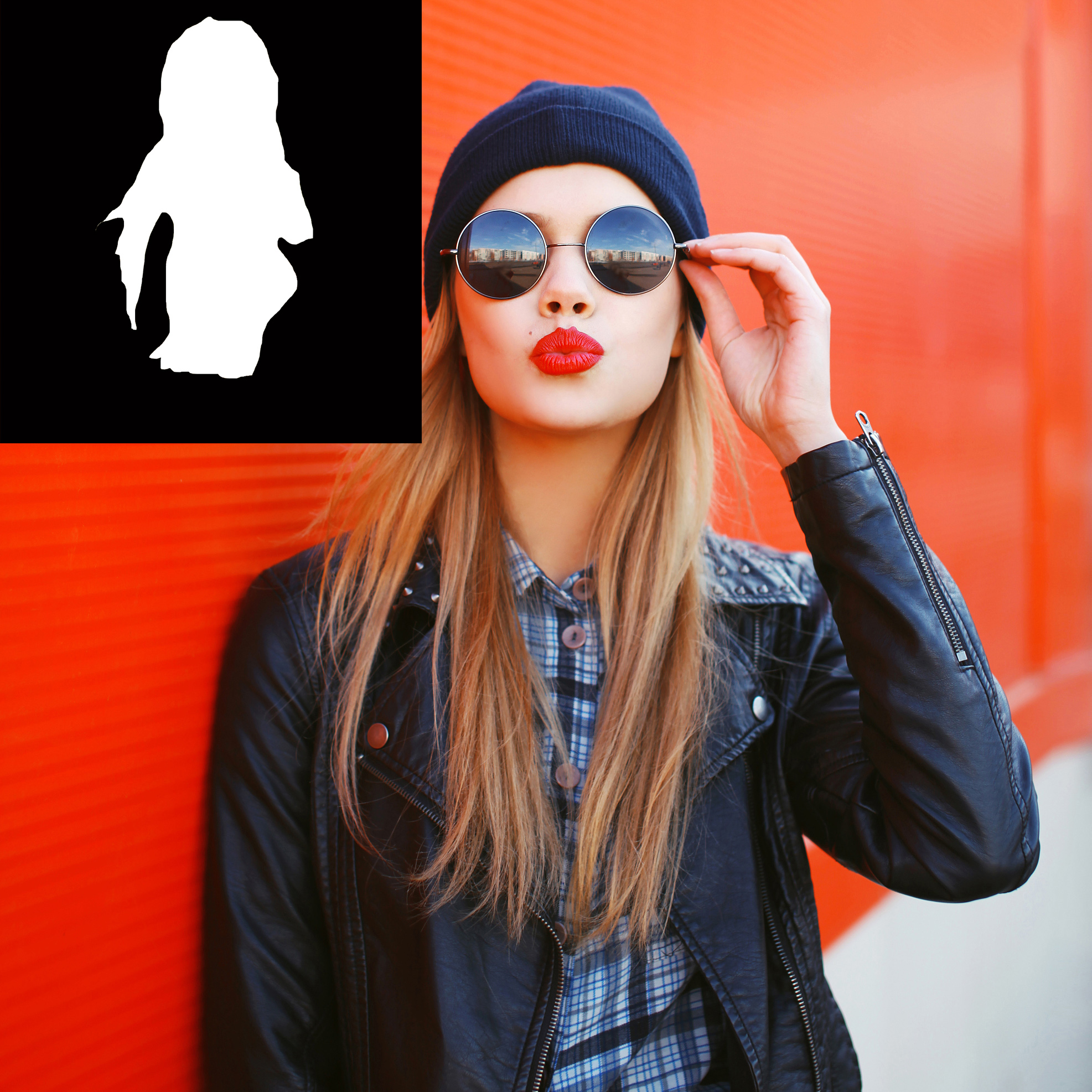}%
\label{ST:Fig:LocalAdjustments:ContentMask}%
}\hfill
\subfloat[Global Transfer and Style]{%
\includegraphics[width=0.195\linewidth]{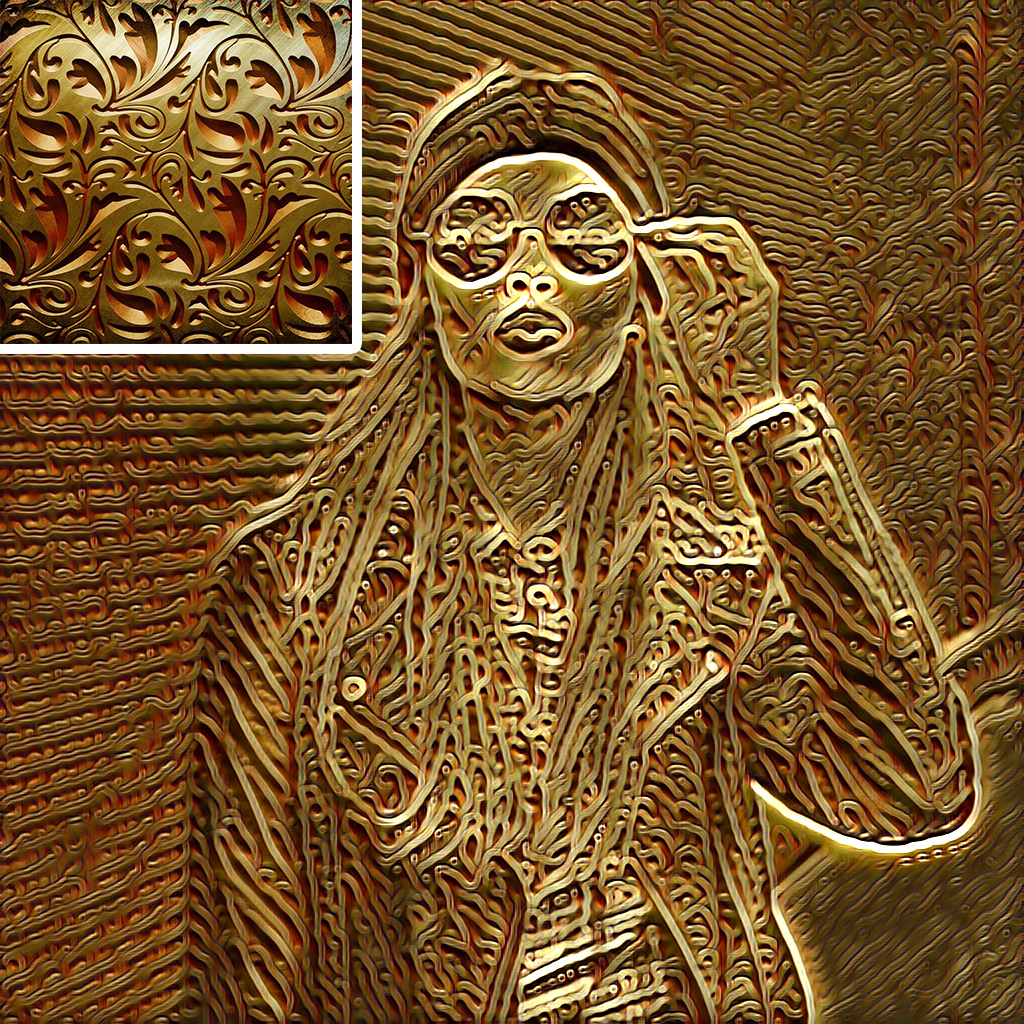}%
\label{ST:Fig:LocalAdjustments:Gold:Global}%
}\hfill
\subfloat[Local Adjustments]{%
\includegraphics[width=0.195\linewidth]{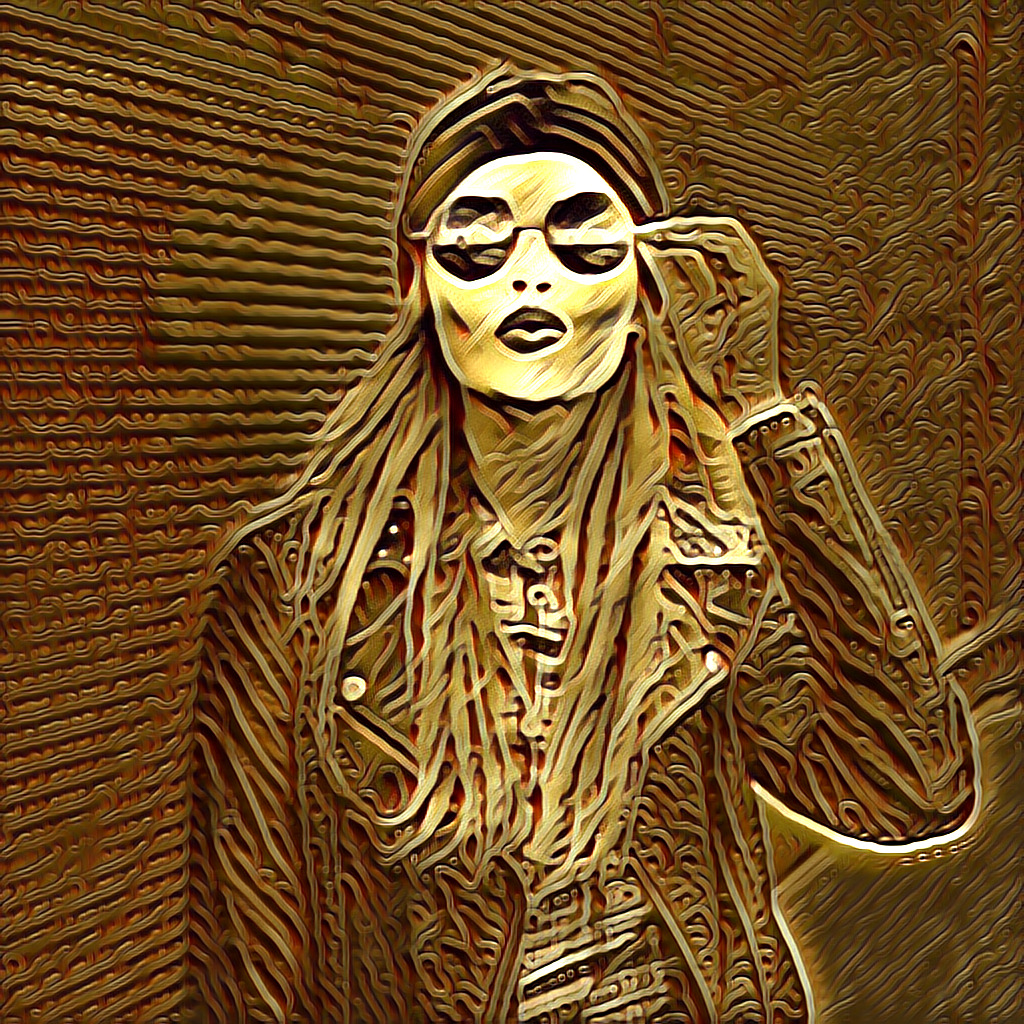}%
\label{ST:Fig:LocalAdjustments:Gold:Local}%
}\hfill
\subfloat[Global Transfer and Style]{%
\includegraphics[width=0.195\linewidth]{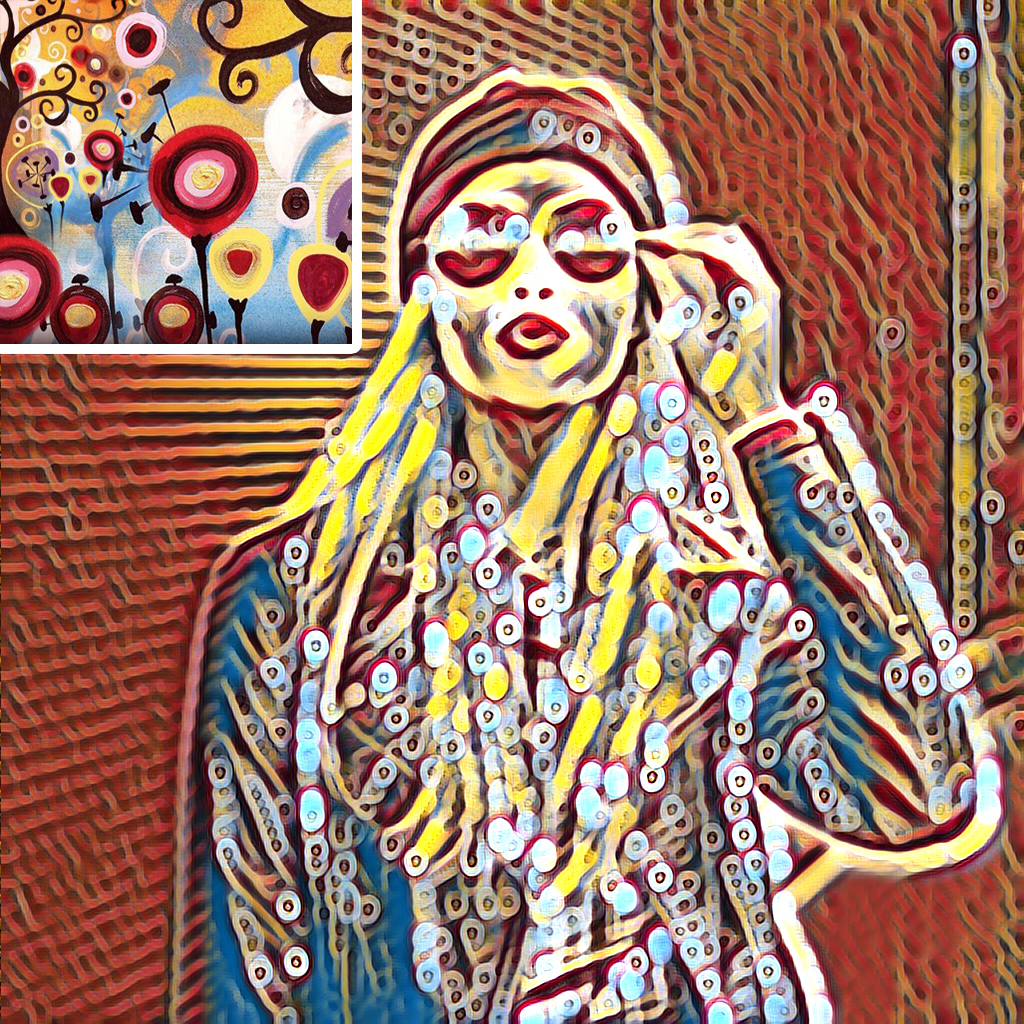}%
\label{ST:Fig:LocalAdjustments:Candy:Global}%
}\hfill
\subfloat[Local Adjustments]{%
\includegraphics[width=0.195\linewidth]{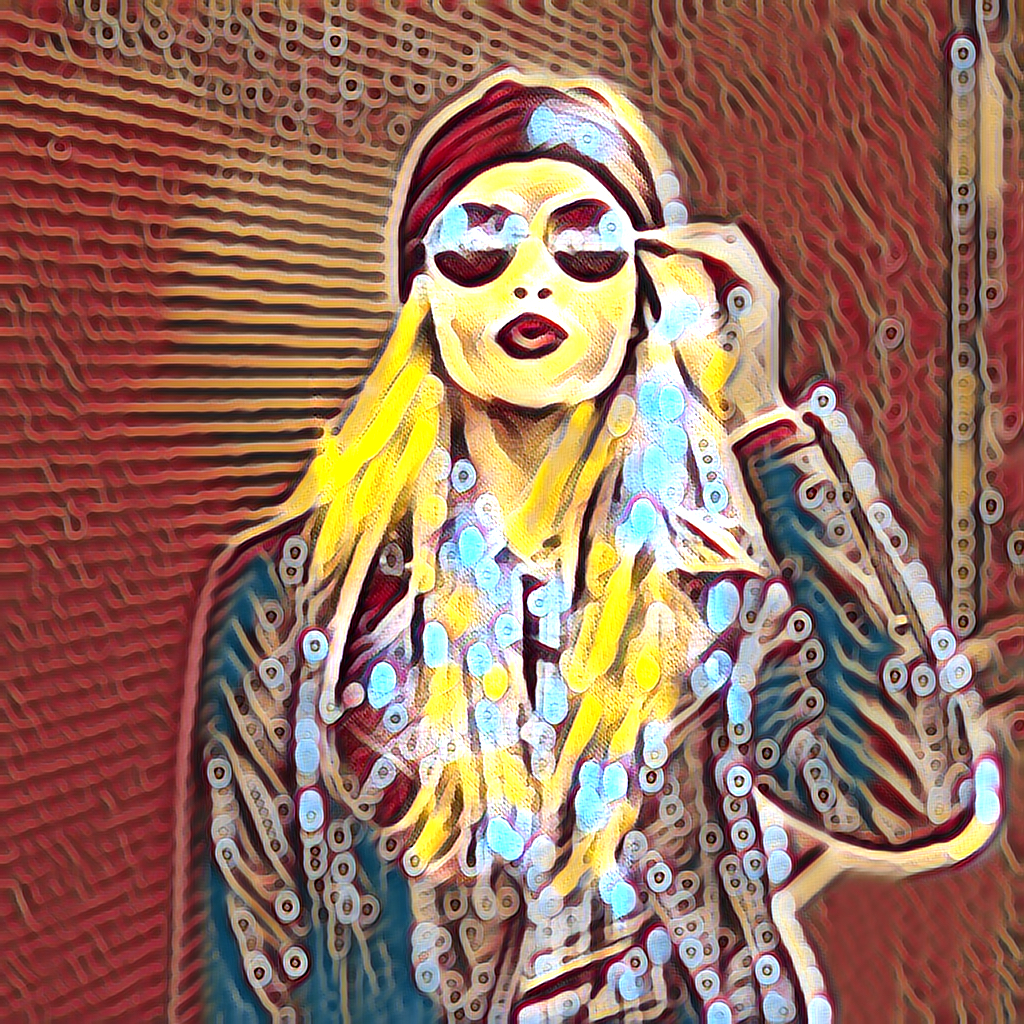}%
\label{ST:Fig:LocalAdjustments:Candy:Local}%
}}
\caption{Comparison between global style transfer and locally retouched versions produced using \StyleTune. }
\label{ST:Fig:LocalAdjustments}
\belowfig 
\end{figure*}

\noindent \Cref{ST:Fig:LocalAdjustments} shows exemplary results obtained using \StyleTune to locally retouch the hair and face of the person. Using less obstructive lines in the face yield a visually more pleasing result and the lighter color tone directs the gaze to the center of the image. Furthermore, oriented strokes along flow lines of the hair in \Cref{ST:Fig:LocalAdjustments:Gold:Local} improves the visual separation of foreground and background and lends the image more depth. The tools provided via \StyleTune can thus be used to art-direct semiotic aspects of the style with local guidance, a step towards \emph{semiotics-based loss functions} \cite{SID17}.

\begin{figure}[tb]
\centering
\includegraphics[trim={0.1cm 0.1cm 0.1cm 0.1cm},clip,width=0.9\linewidth]{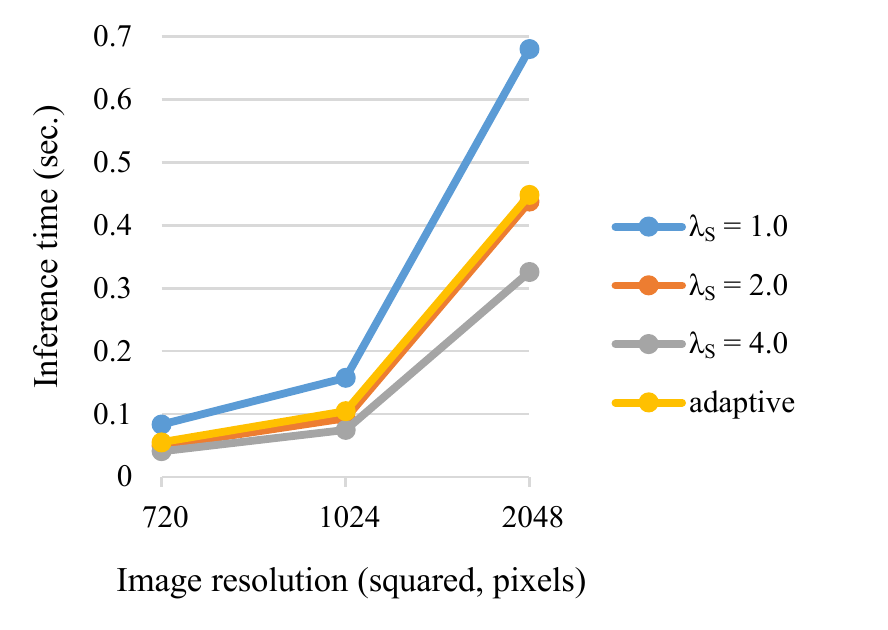}
\caption{Performance comparison for different stroke sizes of our adjustable architecture and the adaptive network \cite{Jing2018}. Tests were performed on desktop PC using a Nvidia GTX 1080Ti GPU and averaged over \num{100} processed images.}
\label{ST:Fig:performance_comparison}
\belowfig 
\end{figure}

\subsection{Performance Considerations}
\label{ST:SubSection:PerformanceEvaluation}

\noindent The performance of the proposed architecture depends mainly on the stroke-size setting, as larger strokes are generated by running through a downsampled style branch.
As \Cref{ST:Fig:performance_comparison} shows, for stroke sizes $\lambda_S \ge 2.0$, our network is faster than the adaptive network of Jing \etal\cite{Jing2018} and the similarly performing network of Johnson \etal\cite{JohnsonAF16}. For better comparability to the performance of other \ac{NST} methods, the timings are shown for a desktop system similar to  \cite{jing2019neural}.

For execution on mobile devices, we tested our application using an iPad Pro \nth{3} generation equipped with an Apple A12X Bionic and \SI{4}{\giga\byte} \ac{RAM}. We use input images of $1024 \times 1024$ pixels resolution. The application of global style transfer and the image-based blending operation run in real-time without notable latency. The style-encoder network for stroke-size pre-generation required approx. \SI{5}{\sec} for \num{10} stroke sizes and the decoder network performing model-space blending approx. \SI{3}{\sec}. A lower number of pre-computed strokes can be sufficient for many styles when editing locally, however for adjusting the global granularity and orientation sliders, \num{10} levels and more empirically provide a visually smooth transition between scales.  The optional server-based upsampling requires \SI{1}{\min} for \num{256} \ac{Mpix}. The overall on-device memory usage comprises approx. \SI{900}{\mega\byte}.

\beforesubsec
\subsection{Limitations}
\label{ST:SubSection:Limitations}

\noindent While \StyleTune enables more degrees of artistic freedom for style transfer, there are still some limitations to overcome. Our proposed architecture achieves a greater range of flexibility for global edits, however, strokes from one stroke-size level are generally not consistently placed at the same location in other stroke-size levels. 
This represents a limitation on local editing scenarios where strokes are expected to consistently flow between different stroke sizes. 

To remedy this, we also implement the adaptive stroke architecture of Jing \etal\cite{Jing2018} in \StyleTune as an option for fine detail control. Further, before locally applying edits, there is a brush pre-generation step for every stroke size, which incurs a loading time that is dependent on the number of brush sizes and any stroke orientation changes require a new round of pre-computation for stroke size previews. Finally, the patch-based upsampling step alters global appearance in ways that may not be intended by the user and has a high execution time.

\section{Conclusions \& Future Work}
\label{ST:Sec:Conclusions}


\noindent This paper introduces an approach for multivariate control over fast style transfer. Our method is the first to enable control over stroke size, style intensity and stroke orientation with a single model. We demonstrate the real-world applicability of our idea by implementing a mobile app for finegrained global and local control over these aspects using our proposed network. Our app implements an editing pipeline to enable both interactive adjusting and retouching of results as well as very high resolution exports using style-guided upsampling. Our work is a step towards making NST a useful tool for art-directed image stylization for casual and professional users, however, still some limitations remain to be addressed.

As future work, we plan to further explore the integration of different style scales into patch-based upsampling by adding a neural representation of stroke size and implement the upsampling method on mobile devices directly, to provide interactive feedback by continuously updating the visual results during editing.

\section*{Acknowledgment}
\noindent This work was partially funded by the \ac{BMBF} through grants 01IS18092 (\enquote{mdViPro}) and 01IS19006 (\enquote{KI-LAB-ITSE}).

\nocite{*}
\bibliographystyle{IEEEtran}
\bibliography{IEEEabrv,styletune.bib}

\begin{acronym}
\acro{BMBF}{German Federal Ministry of Education and Research}
\acro{AABB}{Axis-aligned Bounding Box}
\acro{OS}{Operating System}
\acro{EC2}{Elastic Compute Cloud}
\acro{API}{Application Programming Interface}
\acro{B2C}{Business-to-Customer}
\acro{B2B}{Business-to-Business}
\acro{GCM}{Google Cloud Messaging}
\acro{CPU}{Central Processing Unit}
\acro{GPU}{Graphics Processing Unit}
\acro{HTML}{Hyper Text Markup Language}
\acro{WebGL}{Web Graphics Library}
\acro{RAM}{Random Access Memory}
\acro{VRAM}{Video Random Access Memory}
\acro{UGC}{User-Generated Content}
\acro{OpenGL}{Open Graphics Library}
\acro{REST}{Representational State Transfer}
\acro{vCPU}{virtual CPU}
\acro{NST}{Neural Style Transfer}
\acro{HD}{High Definition}
\acro{FHD}{Full High Definition}
\acro{QHD}{Quad High Definition}
\acro{DRM}{Digital Rights Management}
\acro{UI}{User Interface}
\acro{URL}{Unified Resource Locator}
\acro{W3C}{World Wide Web Consortium}
\acro{OGC}{Open Geospatial Consortium}
\acro{HDR}{High Dynamic Range}
\acro{JPEG}{Joint Photographic Experts Group}
\acro{DOM}{Document Object Model}
\acro{ML}{Machine Learning}
\acro{CSS}{Cascading Style Sheets}
\acro{CNN}{Convolutional Neural Network}
\acro{NSFW}{Not Safe For Work}
\acro{CMS}{Content Moderation Service}
\acro{CAS}{Content Analysis Service}
\acro{IAS}{Image Abstraction Service}
\acro{TOS}{Terms of Service}
\acro{JS}{JavaScript}
\acro{SSD}{Single Shot Detector}
\acro{Mpix}{Megapixel}
\acro{CIN}{Conditional Instance Normalization}
\acro{ONNX}{Open Neural Network Exchange}
\acro{adaIN}{adaptive Instance Normalization}

\end{acronym}

\end{document}